\documentclass[runningheads]{llncs}

\usepackage{eccv}

\usepackage{eccvabbrv}

\usepackage{booktabs}
\usepackage{amsmath}
\usepackage{amssymb}
\usepackage{multirow}
\usepackage{array}
\usepackage{makecell}
\usepackage{amsfonts}
\usepackage{algorithm}
\usepackage{algpseudocode}
\usepackage{bbm}
\usepackage{bm}
\usepackage{xcolor}
\usepackage{pifont}
\usepackage{wrapfig}
\usepackage{dsfont}
\usepackage{cuted}
\usepackage{capt-of}
\usepackage{microtype}
\usepackage{ragged2e}
\usepackage{ulem}
\usepackage{bm}
\usepackage{enumitem}

\usepackage{colortbl}

\definecolor{c_muns}{rgb}{0.88,1,1}
\definecolor{c_msup}{rgb}{1.0,0.88,1}
\definecolor{c_data}{rgb}{0.9,0.9,0.9}
\definecolor{c_lowbest}{rgb}{1.0,0.8,0.8}
\definecolor{c_highbest}{rgb}{0.8,0.8,1.0}
\usepackage[accsupp]{axessibility}  
\DeclareMathAlphabet{\mathpzc}{OT1}{pzc}{m}{it}

\usepackage{hyperref}

\usepackage{orcidlink}

\begin{document}
\title{
ProSR: Semantic-Prototype-Guided Discrete Modeling for
Physically Consistent SAR Super-Resolution \vspace{-0.5cm}
}
\titlerunning{ProSR: Prototype-Guided Super-Resolution}

\author{
Byoungwoo Kim\orcidlink{0009-0003-6261-7161} \and
Munchurl Kim\orcidlink{0000-0003-0146-5419} \vspace{-0.2cm}
}

\authorrunning{B. Kim and M. Kim}

\institute{
Korea Advanced Institute of Science and Technology, Daejeon, Republic of Korea
\email{\{quddn826, mkimee\}@kaist.ac.kr}
}


\maketitle

\begin{figure*}[h]
    \vspace{-4mm}
    \centering
    \includegraphics[width=1.0\linewidth, height=6cm, keepaspectratio]{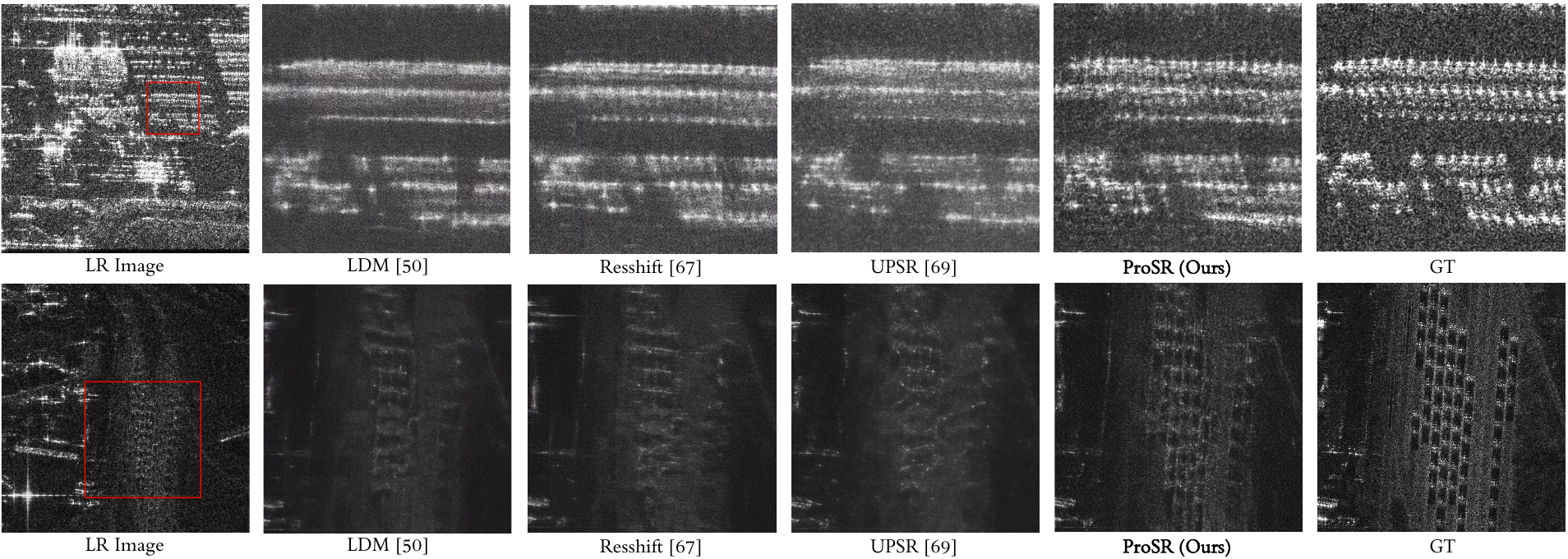}
    \vspace{-6mm}
    \caption{\textbf{Qualitative comparison on SAR ISR ($\times 4$) results on small objects.} Our ProSR restores a structural integrity of tiny targets (red boxes), whereas existing diffusion methods suffer from distortion or blurring. (Zoom in for better view.)}
    \label{fig:first_page_comp}
    \vspace{-10mm}
\end{figure*}
\begin{abstract}
High-resolution Synthetic Aperture Radar (SAR) imagery is critical for precision analysis such as automatic target recognition, yet its acquisition is costly. Although generative image super-resolution (ISR) models offer a promising alternative, current smooth-approximation-based diffusion frameworks often struggle to preserve the coherent scattering statistics, causing stochastic structural distortions that are less consistent with real SAR physics. To address this, we propose Semantic-Prototype-Guided Super-Resolution (ProSR), reformulating SAR ISR as \textit{a semantically-guided discrete token prediction task} within a quantized latent space. By mapping signal features to discrete scattering primitives, ProSR preserves the impulsive nature of SAR without over-smoothing. Furthermore, we integrate a Self-Supervised Learning backbone into SAR ISR to extract label-free semantic priors, overcoming label scarcity. Guided by these priors, we introduce Semantic-Aligned Detail Encoding to decouple high-frequency signals into discrete scattering primitives. In parallel, the Semantic Prototype Map Generator explicitly constructs semantic prototype maps, allowing Prototype-Map-Guided Attention to route the information flows within identical categories and mitigate inter-class interference. To validate our approach, we present a large-scale $0.25\,\text{m}$ resolution benchmark from the Umbra Open Dataset. Experimental results show ProSR achieves superior visual quality while preserving essential scattering characteristics required for practical SAR applications.
\end{abstract}

\section{Introduction}
\label{sec:intro}
\begin{figure}[t]
    \centering
    \includegraphics[width=1\linewidth]{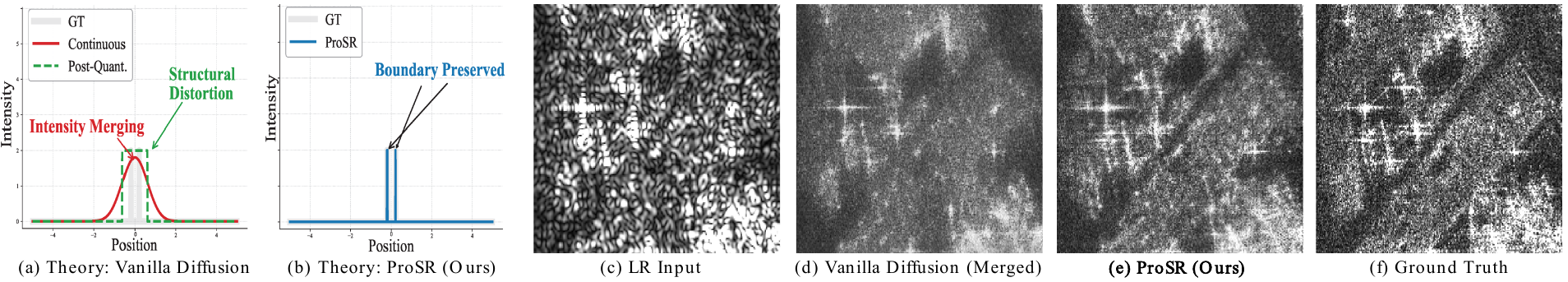}
    \caption{\textbf{Smooth-approximation-based vs. semantic-category-based distribution learning.} (\textit{a, d}) vanilla diffusion models can exhibit stochastic structure distortion (e.g., intensity merging) by averaging neighboring signals. (\textit{b, e}) ProSR (Ours) preserves distinct scattering peaks via predicting semantic-category-based representation.}
    \label{fig:motivation}
\end{figure}
High-resolution (HR) Synthetic Aperture Radar (SAR) imagery is vital for high-precision analysis---such as automatic target recognition (ATR) \cite{zhou2025fifty} and infrastructure monitoring \cite{kyriou2023review}---offering rich scattering information derived from unique interactions between electromagnetic wave and terrain geometry \cite{datcu2023explainable}. However, acquiring such high-fidelity data is costly, due to sensor costs and orbital constraints. Consequently, image super-resolution (ISR) has emerged as a powerful framework to bridge the gap between accessible low-resolution (LR) data and the high-fidelity requirements of practical applications \cite{awais2026classification, bhattacharjee2025hybrid, wang2025metric}.

Generative models, led by smooth-approximation-based diffusion processes, have demonstrated exceptional capabilities in synthesizing sharp, high-resolution natural images \cite{sohl2015deep, ho2020denoising, song2020denoising, rombach2022high, yue2023resshift, zhang2025uncertainty}. However, when applied to the SAR domain, these models often struggle to preserve SAR physical integrity, suffering from what we term \textit{`stochastic structural distortions'}. This degradation can be understood from mode interpolation \cite{aithal2024understanding}—a phenomenon where smooth transitions between disjoint data modes generate samples entirely outside the true data distribution. Fig.~\ref{fig:motivation} compares a smooth-approximation-based versus a semantic-category-based distribution learning and their effects on generated SAR ISR results. The vanilla diffusion models \cite{rombach2022high, yue2023resshift, zhang2025uncertainty} frequently struggle to separate edges that are very close together as shown in Fig.~\ref{fig:motivation}-(a). The two nearby edges (gray color) are difficult to be modeled by the vanilla diffusion models that can yield an overlapped and blurred edge (red color), creating one single fake intensity between the true scattering peaks after post quantization (green color). This structural degradation can be related to the learned score function being a smooth approximation of the data manifold \cite{song2020score, salimans2022progressive}, which often makes it difficult to maintain the sharp discontinuities essential to SAR physics (Fig.~\ref{fig:motivation}-(d)). A detailed analysis of a toy example regarding unresolved conditional scatterer modes is provided in the \textit{suppl. C} to offer intuition for this behavior. To address this problem, we propose Semantic-Prototype-Guided Super-Resolution (ProSR). ProSR reformulates SAR ISR as \textit{a semantically-guided discrete token prediction task}, implemented via Prototype-Map-Guided Masked Generative Modeling (PMG). By leveraging a discrete latent space to map features into a finite set of physically valid codebook entries \cite{esser2021taming}, PMG treats reconstruction as an iterative selection of categorical tokens rather than smooth density approximation. Unlike vanilla diffusion models, as illustrated in Fig.~\ref{fig:motivation}-(a), that tend to average out complex signal distributions, two nearby distinct edges can be faithfully generated, as shown in Fig.~\ref{fig:motivation}-(b), by quantized representation of our ProSR that enforces a strict reconstruction of SAR-specific scattering patterns \cite{datcu2023explainable}. By doing so, the generation process is inherently prevented from generating ambiguous and smeared scattering signals, which is shown in Fig.~\ref{fig:motivation}-(e) unlike the result shown in Fig.~\ref{fig:motivation}-(d). 

Furthermore, while recent trends in SAR ISR \cite{wang2025metric, awais2026classification, bhattacharjee2025hybrid} aim to improve structural fidelity by incorporating task-oriented objectives (e.g., detection losses), these approaches are limited by the scarcity of labeled SAR data. To overcome this label dependency, we introduce a self-supervised representation framework to the SAR ISR task, leveraging foundational models \cite{caron2021emerging, muzeau2024safe} for label-free semantic guidance. This approach allows us (i) to establish Semantic-Aligned Detail Encoding (SADE), which maps discrete codes to specific scattering classes via a semantic-aligned detail codebook, and (ii) to define a semantic-aware perceptual loss.
While these components provide a robust semantic foundation, they cannot inherently prevent contextual leakage or semantic drift during the cross-attention process. To address this, we introduce Semantic Prototype Map Generator (SPMG) and Prototype-Map-Guided Attention (PMGA). Specifically, within PMG, the SPMG constructs spatial-semantic maps that guide PMGA to restrict attention flows to semantically consistent regions. By acting as a spatial-semantic guidance, PMGA eliminates inter-class signal leakage, ensuring that the reconstruction is both physically accurate and aligned with the actual structure. Finally, we provide a $0.25\,\text{m}$ slant range resolution SAR ISR benchmark from the Umbra Open Dataset \cite{umbra2023open} as a public resource to mitigate data scarcity. In summary, the primary contributions of our work are as follows:
\begin{itemize}
\item To the best of our knowledge, ProSR is the first discrete generative framework tailored for SAR ISR. It utilizes PMG to suppresses \textit{stochastic structural distortions} by mapping signal features into a finite set of physically valid codebook entries, ensuring high-frequency fidelity.
\item We establish a label-free semantic guidance strategy by integrating a self-supervised representation model. This enables SADE, SPMG, and PMGA to achieve precise structural anchoring and suppress inter-class interference without requiring scarce manual annotations.
\item We establish a large-scale $0.25\,\text{m}$ resolution SAR ISR benchmark from the Umbra Open Dataset \cite{umbra2023open} enabling robust evaluation and future research.
\item Our ProSR establishes a new state-of-the-art in SAR ISR, demonstrating superior visual realism while better preserving the essential scattering characteristics required for practical SAR applications.
\end{itemize}
\section{Related Works}
\label{sec:related}
\subsection{Single Image Super-Resolution (SISR)}
SISR has transitioned from traditional optimization to deep generative models \cite{su2025review}. While early pixel-wise architectures \cite{dong2015image, kim2016accurate, lim2017enhanced, zhang2018image} often produced over-smoothed results, GAN-based models \cite{ledig2017photo, wang2018esrgan, liang2021swinir, ma2020structure} significantly improved sharpness but frequently introduced unrealistic hallucinations due to adversarial instability. Consequently, recent research has shifted toward diffusion models \cite{saharia2022image, li2022srdiff, yue2023resshift, wang2024exploiting, wu2024seesr, xiao2024semantic, chen2025srsr, liu2025semantic, zhang2025uncertainty} for higher perceptual fidelity and training stability; however, these diffusion frameworks \textit{inherently rely on a smooth approximation of the data distribution} \cite{aithal2024understanding, song2020score, salimans2022progressive}. This induces \textit{mode interpolation} \cite{aithal2024understanding}, where the model generates samples that smoothly bridge disjoint data modes, resulting in structural hallucinations. Although such artifacts may appear plausible in natural images, they degrade the physical integrity required for high-precision SAR image analysis.
\subsection{Discrete Latent Representations}
Discrete representation learning, popularized by VQ-VAE \cite{van2017neural} and VQ-GAN \cite{esser2021taming}, maps high-dimensional data into a finite set of quantized codebook entries. Unlike continuous models, these formulations replace latent values with a categorical selection from a fixed vocabulary. By reformulating reconstruction as a discrete token prediction paradigm \cite{chen2020generative, austin2021structured, chang2022maskgit, chang2023muse, peng2024confidence}, researchers utilize quantization as a formative bottleneck. While lossy for natural images, this bottleneck uniquely favors SAR, precluding blurred intermediate values and forcing the model to select from distinct entries. This mechanism suppresses the averaging effects common in continuous spaces. Leveraging these advantages, we propose ProSR, which incorporates a scattering primitive guidance within the discrete bottleneck. This enables our ProSR to achieve realistic reconstruction while preserving physical scattering distributions, overcoming the blurring artifacts of continuous spaces.
\subsection{SAR Image Super-Resolution (SAR ISR)}
SAR ISR is uniquely challenging due to the presence of speckle noise and complex electromagnetic interactions with geometric structures of the ground. Recent works have moved from mathematical priors \cite{karakucs2020solving, karimi2021convex} to generative paradigms \cite{zhang2023blind, jiang2024lightweight, chen2024super, chen2025dadsr} to synthesize high-frequency signatures. These realistic reconstructions improve downstream performance (e.g., ATR), proving that realistic scattering patterns are vital for operational analysis. To further enhance fidelity, task-driven SAR ISR incorporates auxiliary structural priors such as edge maps or Electro-Optical imagery \cite{yanshan2022ogsrn, zhu2025sar}, or downstream losses such as a detection loss \cite{awais2026classification, bhattacharjee2025hybrid, wang2025metric}. Despite their success, these methods are heavily bottlenecked by the scarcity of labeled SAR datasets. This label dependency significantly limits the generalizability and scalability of task-driven models. Unlike these label-dependent methods, our ProSR overcomes the scarcity of labeled data by leveraging a self-supervised representation model as a semantic anchor.
\subsection{Self-supervised Representation Learning in SAR}
To overcome label scarcity, Self-Supervised Learning (SSL) has become a prevailing approach for learning robust representations from unlabeled data \cite{chen2020simple, he2020momentum, ren2025comprehensive}. Although frameworks like MAE \cite{he2022masked} and DINO \cite{caron2021emerging} have shown that rich semantic features can be extracted from vast unlabeled imagery, their application in SAR has remained largely focused on high-level tasks such as classification or semantic segmentation \cite{li2025saratr, muzeau2024safe}. The potential of SSL-derived semantic priors to serve as a guidance mechanism for low-level reconstruction remains largely untapped. Our ProSR bridges this gap by integrating SSL-derived semantic priors into the reconstruction pipeline. Specifically, we construct a detail codebook aligned with the semantic features of LR inputs and employ a semantic prototype map to guide cross-attention. This enables input-consistent detail recovery while ensuring physical consistency without the need for manual labels.
\begin{figure*}[t]
    \centering
    \includegraphics[width=\linewidth]{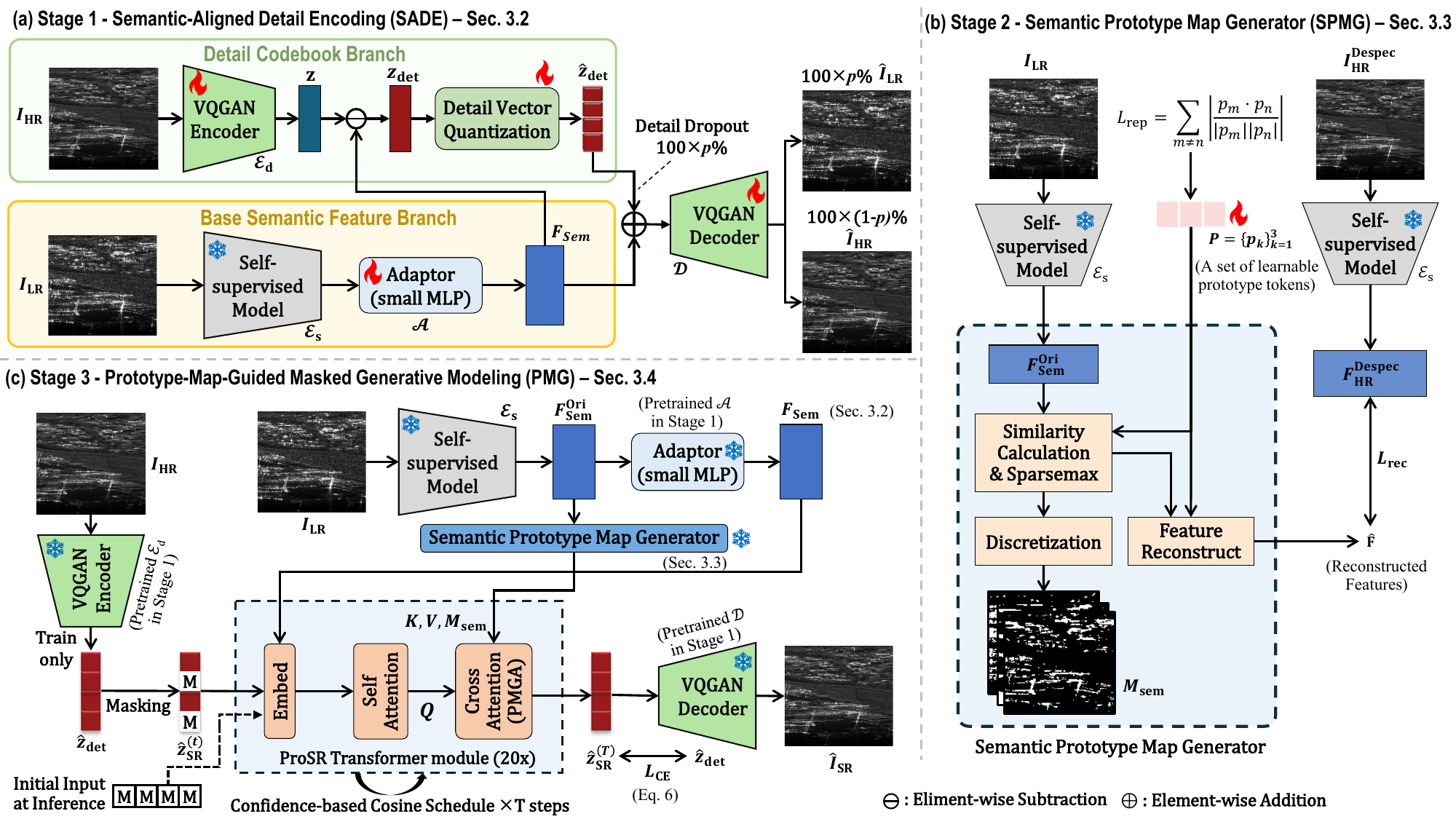}
    \caption{Overall architecture of the proposed ProSR framework.}
    \label{fig:method}
\end{figure*}

\section{Methodology}
\label{sec:method}
\subsection{Overview of ProSR}
Fig.~\ref{fig:method} illustrates the overall architecture of our ProSR framework, which bridges continuous semantic priors with discrete high-frequency details using a spatial-semantic map to address SAR stochastic structural distortions. The framework consists of three stages: (i) Stage 1 - Semantic-Aligned Detail Encoding (SADE), (ii) Stage 2 - Semantic Prototype Map Generation (SPMG), and (iii) Stage 3 - Prototype-Guided Masked Generative Modeling (PMG). It should be noted that Stage 1 - SADE is first \textit{pretrained} with HR SAR ground-truth (GT) images and their LR versions to learn separated representations of details (high-frequency) via the VQGAN Encoder (denoted as $\mathcal{E}_\text{d}$) \cite{esser2021taming} and the Detail Vector Quantization module, and structural information (low-frequency) via a self-supervised model \cite{muzeau2024safe}(denoted as $\mathcal{E}_\text{s}$) and an Adaptor (denoted as $\mathcal{A}$) \cite{chen2025aligning}, where both representations are jointly optimized in conjunction with a shared VQGAN Decoder (denoted as $\mathcal{D}$) \cite{esser2021taming}. Then, the pretrained $\mathcal{E}_s$ is incorporated in Stage 2 for training and inference. In Stage 3, the pretrained $\mathcal{E}_\text{d}$, $\mathcal{E}_\text{s}$ and $\mathcal{A}$ are used for training while the pretrained $\mathcal{E}_\text{s}$, $\mathcal{A}$ and $\mathcal{D}$ are for inference.
\subsection{Stage 1 - Semantic-Aligned Detail Encoding (SADE)}
SADE isolates the high-frequency SAR details of HR (GT) $I_\text{HR} \in \mathbb{R}^{H \times W}$ into codebook representations where $H$ and $W$ are the height and width for $I_\text{HR}$, respectively, while maintaining strict alignment with LR input $I_\text{LR} \in \mathbb{R}^{H/4 \times W/4}$ which is upscaled to the same size of $I_\text{HR}$, as shown in Fig.~\ref{fig:method}-(a). To isolate such high-frequency SAR image details, we employ a dual-encoder architecture. We extract raw semantic features $\mathbf{F}_\text{sem}^\text{Ori} = \mathcal{E}_\text{s}(I_\text{LR}) \in \mathbb{R}^{192 \times (H/8 \cdot W/8)}$ for $I_\text{LR}$. For this, we used the SSL backbone \cite{muzeau2024safe} as $\mathcal{E}_\text{s}$ that was pretrained with our SAR training data. $\mathbf{F}_\text{sem}^\text{Ori}$ is then projected into a lower-dimensional latent space as $\mathbf{F}_\text{sem} = \mathcal{A}(\mathbf{F}_\text{sem}^\text{Ori}) \in \mathbb{R}^{32 \times (H/8 \cdot W/8)}$ via a lightweight adapter $\mathcal{A}$ \cite{chen2025aligning}. While $\mathbf{F}_\text{sem}$ serves as a compact semantic anchor for detail decomposition, $\mathbf{F}_\text{sem}^\text{Ori}$ can be used as high-capacity keys and values for the cross-attention mechanism in Stage 3. 

Simultaneously, $\mathcal{E}_\text{d}$ as a detail encoder maps $I_\text{HR}$ into a latent representation $\bm{z} = \mathcal{E}_\text{d}(I_\text{HR}) \in \mathbb{R}^{32 \times (H/8 \cdot W/8)}$. We define the latent detail feature $\bm{z}_\text{det}$ as the residual between $\bm{z}$ and $\mathbf{F}_\text{sem}$, which is denoted as $\bm{z}_\text{det} = \bm{z} - \mathbf{F}_\text{sem}$. This residual is then quantized into $\hat{\bm{z}}_\text{det}$ which then becomes a codebook entry. By doing so, we restrict the codebook to high-frequency details, effectively decoupling them from spatial semantic context (low-frequency components).

To ensure the isolation of high-frequency details of $I_\text{HR}$, we employ a stochastic training strategy. With a probability of $p$, $I_\text{LR}=\mathcal{D}(\mathbf{F}_\text{sem})$ is reconstructed by compelling the decoder $\mathcal{D}$ to distill the semantic structure from $\mathbf{F}_\text{sem}$ independently. With the remaining $1-p$ probability, $\mathcal{D}$ integrates $\hat{\bm{z}}_\text{det}$ with $\mathbf{F}_\text{sem}$ to synthesize $I_\text{HR}$. The reconstructed outputs for the two branches are obtained as:
\begin{equation}
\hat{I} = 
\begin{cases} 
\hat{I}_\text{LR} = \mathcal{D}(\mathbf{F}_\text{sem}) & \text{with }  p \\ 
\hat{I}_\text{HR} = \mathcal{D}(\hat{\bm{z}}_\text{det} + \mathbf{F}_\text{sem}) & \text{with } 1-p
\end{cases}
\end{equation}
\noindent where $p$ is empirically set to 0.25. This stochastic strategy prevents the detail codebook from encoding redundant structures, isolating pure details, by forcing $\mathbf{F}_\text{sem}$ to maintain a self-sufficient structural representation. Consequently, $\hat{\bm{z}}_\text{det}$ focuses on semantic-aligned scattering signatures (high-frequency) of $I_\text{HR}$ while $\mathbf{F}_\text{sem}$ acts as an independent structural anchor that dictates semantic structures (low-frequency) of $I_\text{LR}$, making $\hat{\bm{z}}_\text{det}$ inherently predictable from $\mathbf{F}_\text{sem}$ in Stage 3.

\noindent \textbf{Gumbel Vector Quantization.} We employ Gumbel Vector Quantization (Gumbel-VQ) \cite{jang2016categorical, esser2021taming} to map $\bm{z}_\text{det}$ into a discrete codebook entry, ensuring stable and differentiable training. To prevent codebook collapse and ensure its uniform utilization, we adopt the KL regularization $\mathcal{L}_\text{KL} = \mathbb{E} \left[ \sum_{k=1}^K \pi_{k} \log(\pi_{k} \cdot K) \right]$, where $\pi_k = P(q(\bm{z}_\text{det}) = \bm{e}_k)$ denotes the probability of $\bm{z}_\text{det}$ being assigned to the $k$-th entry (codebook assignment) $\bm{e}_k$ via the quantizer $q(\cdot)$. This objective maximizes codebook entropy, forcing tokens to represent a diverse and comprehensive library of SAR scattering primitives.

\noindent \textbf{Training Objectives.} Stage 1 is optimized using a stochastic objective function that adaptively selects loss terms from the reconstruction branches as:
\begin{equation}
\mathcal{L}_\text{total} = 
\begin{cases} 
\mathcal{L}_\text{rec} + \lambda_\text{per}\mathcal{L}_\text{per} & \text{with } p \\ 
\mathcal{L}_\text{rec} + \lambda_\text{per}\mathcal{L}_\text{per} + \lambda_\text{adv}\mathcal{L}_\text{adv} + \lambda_\text{KL}\mathcal{L}_\text{KL} & \text{with } 1-p 
\end{cases}
\end{equation}
where $\mathcal{L}_\text{rec}$ and $\mathcal{L}_\text{adv}$ \cite{esser2021taming} are an $\mathcal{L}_\text{1}$-reconstruction and adversarial losses, respectively. To capture domain-specific signatures without labeled SAR data, we define the perceptual loss $\mathcal{L}_\text{per}$ as $\mathcal{L}_\text{SAFE}$ using pre-trained a SAR SSL model $\mathcal{E}_\text{s}$ \cite{muzeau2024safe}, enabling physically-grounded supervision from unlabeled imagery. This approach ensures the preservation of essential scattering patterns that pixel-wise losses often overlook. Restricting $\mathcal{L}_\text{adv}$ and $\mathcal{L}_\text{KL}$ to the $1-p$ branch forces the codebook toward realistic high-frequency signatures, while $\mathbf{F}_\text{sem}$ preserves the fundamental semantic structural layout.

\subsection{Stage 2 - Semantic Prototype Map Generation (SPMG)} \label{subsec:spmg}
The SPMG converts continuous semantic features into a discrete spatial-semantic map (Fig.~\ref{fig:method}-b), bridging the encoding-generation gap stemming from ignoring semantic information during cross-attention.

\noindent \textbf{Step 1: Mapping Spatial Features to Semantic Prototype Tokens.}
We define a set of $K=3$ learnable prototype tokens, $P = \{\bm{p}_\text{T}, \bm{p}_\text{S}, \bm{p}_\text{C}\} \in \mathbb{R}^{3 \times C}$, representing the fundamental scattering primitives in the SAR domain: target (T), shadow (S), and clutter (C).
The choice of a token ensures a corresponding semantic interpretability by aligning it with a physical scattering primitive. Empirically, we found that $K > 3$ often leads to redundant or inactive prototype tokens, whereas $K=3$ maintains an effective representation throughout all our experiments. For the semantic feature $\mathbf{F}_\text{Sem}^\text{Ori}$, we denote $\bm{f}_{i} \in \mathbb{R}^C$ as a spatial feature vector at spatial position $i$ of $\mathbf{F}_\text{Sem}^\text{Ori}$. For $\bm{f}_{i}$, we compute a sparse similarity assignment $S_{i,k}$ for a prototype token $\bm{p}_k \in P$ using the Sparsemax \cite{martins2016softmax} as:
\begin{equation}
S_{i,k} = \left[ \operatorname{Sparsemax}(-\|\bm{f}_{i} - \bm{p}_k\|_2^2) \right]_k
\label{eq:sparsemax_assignment}
\end{equation}
\noindent \textbf{Step 2: Semantic Map generation via Dynamic Discretization.} Instead of a naive argmax, which often blurs the physical distinction between target and clutter, we employ dynamic discretization to isolate high-intensity regions as structural anchors, ensuring a physical reliability (see \textit{Suppl. B} for sensitivity analysis). At each spatial position $i$ and for each class $k \in \{\text{T, S}\}$, we define a binary indicator mask $\mathcal{I}_{i,k} \in \{0, 1\}$ as:
\begin{equation}
\mathcal{I}_{i,k} = \mathds{1}(S_{i,k} \geq \tau_k \text{ and } S_{i,k} \geq S_{i,m}), \quad k, m \in \{\text{T, S}\}, \quad m \neq k
\label{eq:indicator_mask}
\end{equation}
where $\tau_{k} = \gamma \cdot \max_{j}(S_{j,k})$ is a relative threshold with $j$ spanning all spatial indices. We empirically set $\gamma = 0.7$ to prioritize high-confidence scattering centers while filtering stochastic artifacts (see \textit{Suppl. B} for sensitivity analysis). To ensure semantic exclusivity, clutter regions (C) are defined as the regions that do not belong to target and shadow regions, satisfying $\sum_{k \in \{\text{T,S}\}} \mathcal{I}_{i,k} = 0$. To isolate stable scattering signatures, we intersect the top $c\%$ global $S_{i,\text{C}}$ pixels with the clutter region where $c$ is empirically set to 30. This sparse filtering, prioritizing the classes with relatively higher similarity scores, yields a non-overlapping Semantic Prototype Map ($M_\text{sem}$) that robustly guides Stage 3 to enable semantic-guided cross-attention while eliminating boundary ambiguity.

\noindent \textbf{Prototype Optimization Objectives.}
To anchor the learnable prototype tokens to physical reality, we reconstruct feature map $\hat{\mathbf{F}} \in \mathbb{R}^{192 \times (H/8 \cdot W/8)}$ using prototype tokens $P$. The reconstructed feature at position $i$ is formulated as $\hat{\bm{f}}_i = \sum_{k \in \{\text{T,S,C}\}} S_{i,k} \cdot \bm{p}_k$, where $S_{i,k}$ is the sparse similarity score from Eq.~\eqref{eq:sparsemax_assignment} and $\bm{p}_k \in P$ is the corresponding learnable prototype token. We optimize $\hat{\mathbf{F}}$ against despeckled HR features $\mathbf{F}_\text{HR}^\text{despec} = \mathcal{E}_\text{s}(I_\text{HR}^\text{despec})$ using a feature reconstruction loss $\mathcal{L}_\text{rec} = \|\mathbf{F}_\text{HR}^\text{despec} - \hat{\mathbf{F}}\|_1$, where $I_\text{HR}^\text{despec}$ is obtained following \cite{dalsasso2021if}. To ensure distinctness between target, shadow, and clutter, we use a Repulsion Loss $\mathcal{L}_\text{rep} = \sum_{m\neq n} \left| \frac{\bm{p}_m \cdot \bm{p}_n}{\|\bm{p}_m\| \|\bm{p}_n\|} \right|$ that enforces the orthogonality between prototype token pairs. The final objective is $\mathcal{L}_\text{total} = \mathcal{L}_\text{rec} + \lambda_\text{rep}\mathcal{L}_\text{rep}$, ensuring that the learnable tokens $P$ are both physically representative and semantically discriminative.

\subsection{Stage 3 - Prototype-Map-Guided Masked Generative Modeling}
Prototype-Map-Guided Masked Generative Modeling (PMG) (Fig.~\ref{fig:method}-c) synthesizes HR tokens using a ProSR Transformer based on the MaskGIT framework \cite{chang2022maskgit}. By reformulating reconstruction \textit{as an iterative selection of categorical tokens} rather than smooth density approximation, our ProSR suppresses \textit{mode interpolation} \cite{aithal2024understanding} and ensures that generated signals remain within physically valid scattering primitives. Input embeddings for the ProSR transformers are formed by concatenating masked detail tokens with $\mathbf{F}_\text{sem}$ from Stage 1, providing a strong semantic prior of $I_\text{LR}$ throughout the iterative process.

\noindent \textbf{Prototype-Map-Guided Attention (PMGA).} 
While self-attention captures long-range spatial dependencies, unconstrained cross-attention between latent tokens and semantic priors ($\mathbf{F}_\text{sem}^\text{Ori}$) can lead to contextual leakage due to the absence of semantic guidance, suffering from semantic drift, where scattering signatures from disparate terrain types may interfere during the cross-attention process. To address this, we introduce PMGA as a spatial-semantic guidance. Using $M_\text{sem}$ (defined in Sec.~\ref{subsec:spmg}), PMGA restricts cross-attention weights $A_{ij} = \operatorname{Softmax} (Q_i K_j^T / \sqrt{d} + M_{ij})$ such that HR tokens only aggregate information from semantically consistent regions. Here, $Q_i$ is the latent query derived from self-attention at position $i$, and $K_j$ is latent key with $\mathbf{F}_\text{sem}^\text{Ori}$ at position $j$. The Semantic Prototype Mask $M_{ij}$ is defined as:
\begin{equation}
M_{ij} = \begin{cases} 0, & \text{if } \mathpzc{l}(\text{pos}_i) = \mathpzc{l}(\text{pos}_j) \\ -\infty, & \text{otherwise} \end{cases}
\end{equation}
where $\mathpzc{l}(\text{pos}_i) \in \{\text{T, S, C}\}$ denotes a semantic label at position $i$ obtained from $M_\text{sem}$. For unassigned regions, cross-attention is bypassed to block background interference. Enforcing intra-class attention suppresses inter-class feature leakage while reinforcing the distinct signatures of individual scattering centers, preserving the sharp intensity distributions and speckle statistics inherent to SAR signals.

\noindent \textbf{Training Objectives.} PMG is optimized by minimizing the cross-entropy (CE) loss between predicted and HR (GT) tokens, conditioned on the visible context and semantic guides:
\begin{equation}
\label{eq:stage2_loss}
\mathcal{L}_{\text{stage3}} = \mathbb{E}_{\hat{\bm{z}}_\text{det}, M} \left[ \sum\nolimits_{i\in M} \mathcal{L}_{\text{CE}} \left( \hat{\bm{z}}_\text{det}^{(i)}, p_\theta( \hat{\bm{z}}_\text{SR}^{(i)} \mid \hat{\bm{z}}_\text{det}^{\bar{M}}, \mathbf{F}_{\text{sem}}^{\text{ori}}, \mathbf{F}_{\text{sem}}, M_{\text{sem}}) \right) \right]
\end{equation}
where $\hat{\bm{z}}_\text{det} = \{\hat{\bm{z}}_\text{det}^{(i)}\}_{i=1}^N$ is the collection of HR tokens, ${\hat{\bm{z}}_\text{det}^{(i)}}$ denotes the token at index $i$, and $p_\theta$ is the likelihood predicted by PMG Transformer $\theta$. The indices are partitioned into a masked set $M$ and a visible set $\bar{M}$, where $\hat{\bm{z}}_\text{det}^{\bar{M}}$ serves as the context for predicting the masked tokens. By minimizing $\mathcal{L}_{\text{stage3}}$ in the discrete latent space, PMG learns to predict high-frequency scattering primitives strictly aligned by the spatial-semantic guidance map $M_{\text{sem}}$. This formulation effectively circumvents the mode interpolation problem, helping the reconstructed intensities remain within the support of physically plausible SAR characteristics.

\noindent \textbf{Inference: Iterative Sampling with PMGA.} At inference, our ProSR synthesizes the HR detail tokens via an iterative decoding process following the MaskGIT framework \cite{chang2022maskgit}. Starting from a fully masked grid, the number of tokens to be unmasked at each step $t$ is determined by a decreasing masking schedule $\gamma(t)$ \cite{chang2022maskgit}. At each iteration, the highest-confidence tokens are fixed while the remainder are re-masked for the next step. Throughout this process, PMGA remains active to ensure that the ProSR Transformer only attends to semantically consistent regions in $\mathbf{F}_\text{sem}^\text{Ori}$. This iterative refinement, guided by $M_{\text{sem}}$, produces high-fidelity SAR image details aligned with the physical scattering layout.

\section{Experimental Results}
\subsection{Dataset Construction}
To evaluate our ProSR under realistic conditions, we curated a high-fidelity SAR dataset from the Umbra Open Dataset \cite{umbra2023open}. We utilized X-band Single Look Complex (SLC) data to serve as the source for HR reference imagery.

\noindent \textbf{HR-LR Pairs.} The source SLC data possesses a high-precision native resolution (${\sim}0.25\,\text{m}$ azimuth, ${\sim}0.25\,\text{m}$ range). To establish a uniform benchmark, we standardized these to $0.25\,\text{m}$ for SAR HR references and $1.0\,\text{m}$ for SAR LR counterparts. Following the sub-aperture decomposition methodology \cite{muzeau2024safe}, the SAR LR references were achieved via spectral domain cropping (sub-sampling the Doppler and range spectra) rather than naive image-space interpolation. This ensures physically grounded resolution degradation while preserving authentic speckle statistics and phase integrity in slant-range geometry. Subsequently, the SAR LR images were oversampled by zero-padding GT spectra in the frequency domain prior to the inverse transform.

\noindent \textbf{Amplitude Extraction and Normalization.} 
Following \cite{debuysere2025quantitative}, we adopted a normalization strategy optimized for generative SAR tasks. To manage the high dynamic range, we scaled the amplitude ($A$) using the scene-level mean ($\mu$) and standard deviation ($\sigma$): $A_\text{norm} = A / (\mu + 3\sigma)$. The normalized values were clipped to $[0, 1]$ to preserve predominant scattering while mitigating extreme outliers, then quantized into 8-bit integers for storage.

\noindent \textbf{Data Filtering and Diversity.} Physical consistency is paramount for model convergence. Since over $97\%$ of our Umbra samples \cite{umbra2023open} are in VV polarization, we constructed a VV SAR dataset to prevent training biases from heterogeneous scattering signatures. We further restricted incidence angles from $10^\circ$ to $50^\circ$, excluding extreme angles due to significant physical divergence and data scarcity.

\noindent \textbf{Preprocessing and Statistics.} The SAR images were tiled into non-overlapping $1024 \times 1024$ patches. To ensure a balanced representation of terrestrial features, we applied a selective quality control filter that prioritized patches with significant scattering structures while retaining a controlled portion of low-signal regions. The curated dataset comprises $132,452$ VV-polarized patches from $502$ unique SAR images, with heights and widths ranging from approximately $4.5\text{K}$ to $66\text{K}$ and $10\text{K}$ to $94\text{K}$ pixels, respectively. These images span a diverse global geographic distribution (See Fig. 6 in \textit{Suppl. A}) to ensure broad environmental representation, with their detailed statistics summarized in Table 6 of \textit{Suppl. A}. More details for the characteristics of SAR images are described in \textit{Suppl. A}.  

\noindent \textbf{Data Availability.}
Our curated dataset is available at \url{https://github.com/KAIST-VICLab/ProSR}.
\subsection{Implementation Details}
ProSR was implemented in PyTorch \cite{paszke2019pytorch} and trained on dual NVIDIA A6000 GPUs using $256 \times 256$ random crops. Data augmentation included horizontal flips and random rotations ($90^\circ$, $180^\circ$, $270^\circ$). For a fair comparison, all baseline SR networks (Table~\ref{tab:full_sr_results}) were trained from scratch on our data without metric-specific early stopping, following their official implementations with SAR-adapted inputs. For AE-based baselines, pretrained AEs remained frozen while only the diffusion ISR networks were trained from scratch. For full-image evaluation, all methods use 256$\times$256 sliding-window inference (stride 128) and weighted averaging.

\noindent \textbf{Stage 1 - SADE.} Stage 1 follows the ResShift \cite{yue2023resshift} architecture ($f=8$) with the encoder ($\mathcal{E}_\text{d}$)’s final projection modified to 32 dimensions. We employed Gumbel-VQ \cite{jang2016categorical, esser2021taming} with 1,024 codebook entries and $32$ embedding dimensions, stabilized by $\lambda_\text{KL}=10^{-6}$. SADE was trained for 500K iterations using AdamW (batch size 12 on a single GPU, learning rate $1 \times 10^{-4}$) \cite{loshchilov2017decoupled} with a composite loss: $\mathcal{L}_\text{total} = \mathcal{L}_\text{rec} + \lambda_\text{SAFE} \mathcal{L}_\text{SAFE} + \lambda_\text{adv} \mathcal{L}_\text{adv}$, where $\lambda_\text{SAFE}$ is set to 0.5 and, $\lambda_\text{adv}$ is adaptively balanced \cite{esser2021taming}. Simultaneously, semantic features were extracted via a SAFE ViT-tiny model $\mathcal{E}_\text{s}$ \cite{muzeau2024safe}, pre-trained on our SAR dataset, and projected to 32 embedding dimensions via a lightweight adapter \cite{chen2025aligning}.

\noindent \textbf{Stage 2 - SPMG.} The embedding space of $P$ was aligned with the ViT-Tiny feature space \cite{dosovitskiy2020image}. The SPMG was trained for 15k iterations (AdamW, LR $1 \times 10^{-4}$, batch size 64, $\lambda_{rep}=2$), with other hyperparameters following Sec.~\ref{subsec:spmg}.

\noindent \textbf{Stage 3 - PMG.} 
PMG Transformer \cite{dosovitskiy2020image} comprises $20$ layers with $8$ attention heads and a $512$-dimensional hidden space. We utilized a Positional Encoding Generator \cite{chu2023CPVT}, facilitating stable sliding-window inference. Training was conducted for 600K iterations using AdamW (LR $1 \times 10^{-4}$, weight decay $0.045$, $\beta_1=0.9, \beta_2=0.96$) with a total batch size 32. After a $10$K-iteration linear warm-up, the learning rate was governed by a Cosine Annealing schedule \cite{loshchilov2016sgdr}, decaying from $1 \times 10^{-4}$ to $1 \times 10^{-7}$. Additionally, a label smoothing factor of $0.1$ was applied to the CE loss to ensure robust token synthesis.

\noindent \textbf{Inference and Sampling.} 
For token synthesis, we followed the MaskGIT \cite{chang2022maskgit} sampling process, employing a cosine schedule $\gamma(t)$ over $8$ steps with a temperature of $4.5$ for logit sampling.

\subsection{Performance Comparison}

\noindent \textbf{Evaluation Metrics.}
To comprehensively evaluate our ProSR with other methods, we used three categories of metrics calculated in the SAR amplitude domain:
\begin{itemize}
    \item \textbf{Standard Fidelity:} Peak Signal-to-Noise Ratio (PSNR) and Structural Similarity Index Measure(SSIM)~\cite{wang2004image} evaluate pixel-level fidelity and local structural agreement, respectively. However, these metrics can favor spatially conservative or over-smoothed reconstructions and therefore do not fully characterize SAR-specific scattering fidelity. Furthermore, SSIM depends strongly on the local covariance ($\sigma_{xy}$) between spatially aligned structures; small scatterer displacements, or merging can reduce this covariance and lower SSIM, in some cases even below that of oversampled LR images. We therefore interpret PSNR and SSIM jointly with complementary SAR-oriented structural, perceptual, and distributional metrics.

    \item \textbf{Target and Structural Integrity:} To evaluate physical and semantic consistency, we prioritize metrics focusing on fine-grained structural fidelity and scattering-sensitive signal integrity. This includes Information-content Weighted SSIM (IW-SSIM) \cite{wang2010information} to assess fidelity in target-dense areas and Haar Perceptual Similarity Index (HaarPSI) \cite{reisenhofer2018haar} for local structural coherence and edge integrity. Additionally, we employ Target-to-Clutter Ratio (TCR)—calculated using prototype-derived target masks—to verify the radiometric fidelity. Rather than simply seeking higher TCR, we evaluate the absolute deviation from the GT ($|\Delta \text{TCR}|$) to ensure the reconstructed scattering intensity remains physically consistent with the original observation.
    
    \item \textbf{Statistical and Perceptual Realism:} Fréchet Inception Distance (FID) \cite{heusel2017gans}, Density (Dens), and Coverage (Cov) \cite{naeem2020reliable} are used to evaluate statistical distributions, quantifying the alignment and overlap between the synthesized and GT data manifolds. Perceptual Realism is measured via Learned Perceptual Image Patch Similarity (LPIPS) \cite{zhang2018unreasonable}, and Deep Image Structure and Texture Similarity (DISTS) \cite{ding2020image} to evaluate structural and textural similarity.
\end{itemize}
\begin{figure}[p] 
    \centering
    \includegraphics[width=1\linewidth, height=0.96\textheight, keepaspectratio]{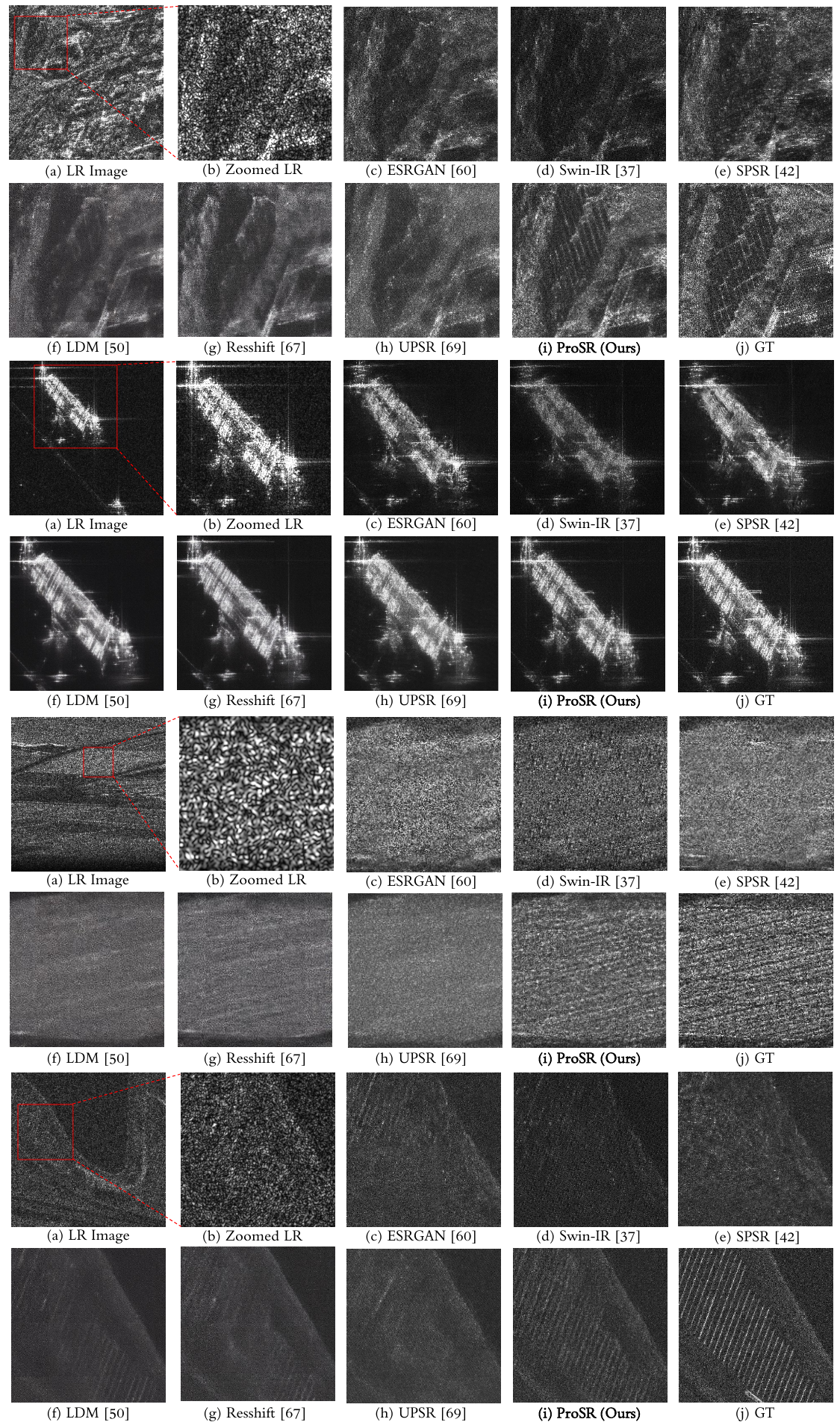}
    \caption{Qualitative comparison of SAR ISR ($\times$4) results (Zoom in for better view)}
    \label{fig:SR_results}
\end{figure}

\begin{table}[t]
\centering
\caption{Quantitative comparison on our SAR ISR ($\times$4) benchmark. The best and second-best results are highlighted in \textbf{\textcolor{red}{bold red}} and \textcolor{blue}{blue}, respectively.}
\label{tab:full_sr_results}
\vspace{-2mm}
\tiny
\resizebox{\textwidth}{!}{%
\begin{tabular}{l|cc|ccc|ccccc}
\toprule
\multirow{2}{*}{Methods} & \multicolumn{2}{c|}{Standard Fidelity} & \multicolumn{3}{c|}{Target \& Structural Integrity} & \multicolumn{5}{c}{Statistical \& Perceptual Realism} \\
 & PSNR $\uparrow$ & SSIM $\uparrow$ & $|\Delta \text{TCR}| \downarrow$ (TCR) & IW-SSIM $\uparrow$ & HaarPSI $\uparrow$ & FID $\downarrow$ & Dens $\uparrow$ & Cov $\uparrow$ & LPIPS $\downarrow$ & DISTS $\downarrow$ \\
\midrule
GT (Reference) & - & - & 0.0 (3.3168) & - & - & - & - & - & - & - \\
\midrule
Oversampled & 16.2056 & \textcolor{red}{\textbf{0.1545}} & 0.6646 (3.9814) & 0.3478 & 0.4211 & 70.86 & 0.2887 & 0.4379 & 0.8062 & 0.4370 \\
\midrule
ESRGAN & 16.1841 & 0.1012 & 0.2223 (3.0945) & 0.3480 & \textcolor{blue}{0.4655} & 50.84 & \textcolor{blue}{0.7249} & 0.6912 & \textcolor{red}{\textbf{0.2977}} & \textcolor{red}{\textbf{0.1372}} \\
SwinIR-GAN & 15.9363 & 0.0959 & 0.5142 (2.8026) & 0.3306 & 0.4407 & 64.73 & 0.5156 & 0.4362 & 0.3071 & 0.2237 \\
SPSR & 16.2756 & 0.1028 & 0.2831 (3.0337) & 0.3451 & 0.4578 & 48.38 & 0.5886 & 0.5936 & 0.3184 & \textcolor{blue}{0.1470} \\
\midrule
LDM-15 & \textcolor{blue}{17.6427} & 0.1277 & 0.0892 (3.2276) & 0.4065 & 0.4017 & 46.91 & 0.4946 & 0.6057 & 0.3888 & 0.2708 \\
ResShift & 17.4784 & 0.1273 & \textcolor{blue}{0.0657} (3.3825) & 0.4066 & 0.4359 & \textcolor{blue}{34.19} & 0.6156 & \textcolor{blue}{0.7022} & 0.3575 & 0.2199 \\
UPSR & \textcolor{red}{\textbf{17.8545}} & \textcolor{blue}{0.1323} & 0.2815 (3.0353) & \textcolor{blue}{0.4201} & 0.4262 & 45.73 & 0.6429 & 0.6665 & 0.3801 & 0.2210 \\
\midrule
\textbf{ProSR (Ours)} & 16.9293 & 0.1088 & \textcolor{red}{\textbf{0.0329}} (3.2839) & \textcolor{red}{\textbf{0.4206}} & \textcolor{red}{\textbf{0.4766}} & \textcolor{red}{\textbf{23.70}} & \textcolor{red}{\textbf{0.9741}} & \textcolor{red}{\textbf{0.8592}} & \textcolor{blue}{0.3010} & 0.1532 \\
\bottomrule
\end{tabular}%
}
\end{table}
\vspace{-2mm}
\begin{table}[t]
    \centering
    \tiny
    \begin{minipage}[t]{0.48\textwidth}
        \centering
        \caption{Stage 1 reconstruction comparison. LDM and ResShift use the same pretrained VQGAN autoencoder \cite{esser2021taming}.}
        \label{tab:ae_comparison}
        \vspace{-2mm}
        \begin{tabular*}{\linewidth}{@{\extracolsep{\fill}} l c c c @{}}
            \toprule
            Models & PSNR $\uparrow$ & SSIM $\uparrow$ & LPIPS $\downarrow$ \\
            \midrule
            LDM/ResShift AE \cite{rombach2022high,yue2023resshift}
            & \textbf{18.3558} & \textbf{0.4072} & \textbf{0.1681} \\
            \midrule
            Ours (1024) & 18.1018 & 0.3799 & 0.2044 \\
            \bottomrule
        \end{tabular*}
    \end{minipage}
    \hfill
    \begin{minipage}[t]{0.48\textwidth}
        \centering
        \caption{Ablation on quantization types.}
        \label{tab:ablation_quant}
        \vspace{-2mm}
        \begin{tabular*}{\linewidth}{@{\extracolsep{\fill}} l c c c @{}}
            \toprule
            Quantization Types & PSNR $\uparrow$ & SSIM $\uparrow$ & LPIPS $\downarrow$ \\
            \midrule
            HR-VQ & 16.5177 & 0.1141 & 0.2765 \\
            Dual-VQ & 17.6195 & 0.3296 & 0.2165 \\
            \textbf{Detail-VQ (Ours)} & \textbf{18.1018} & \textbf{0.3799} & \textbf{0.2044} \\
            \bottomrule
        \end{tabular*}
    \end{minipage}\par\vspace{2mm}\noindent
    \begin{minipage}[t]{0.48\textwidth}
        \centering
        \caption{Ablations on codebook size $K$.}
        \label{tab:codebook_ablation}
        \vspace{-2mm}
        \begin{tabular*}{\linewidth}{@{\extracolsep{\fill}} l c c c @{}}
            \toprule
            Codebook Size $K$ & PSNR $\uparrow$ & SSIM $\uparrow$ & LPIPS $\downarrow$ \\
            \midrule
            512 & 17.9659 & 0.3641 & 0.2124 \\
            \textbf{1024 (Selected)} & \textbf{18.1018} & 0.3799 & 0.2044 \\
            2048 & 18.0894 & \textbf{0.3892} & \textbf{0.2020} \\
            \bottomrule
        \end{tabular*}
    \end{minipage}
    \hfill
    \begin{minipage}[t]{0.48\textwidth}
        \centering
        \caption{Ablation study on the PMGA.}
        \label{tab:ablation_attn}
        \vspace{-2mm}
        \begin{tabular*}{\linewidth}{@{\extracolsep{\fill}} l c c c c @{}}
            \toprule
            Attention Types & IW-SSIM $\uparrow$ & FID $\downarrow$ & Dens $\uparrow$ & Cov $\uparrow$ \\
            \midrule
            Vanilla Cross-Attn. & 0.4186 & 25.39 & 0.9404 & 0.8499 \\
            \textbf{PMGA (Ours)} & \textbf{0.4206} & \textbf{23.70} & \textbf{0.9741} & \textbf{0.8592} \\
            \bottomrule
        \end{tabular*}
    \end{minipage}
\end{table}
\begin{figure*}[t]
    \centering
    \includegraphics[width=1.0\linewidth, keepaspectratio]{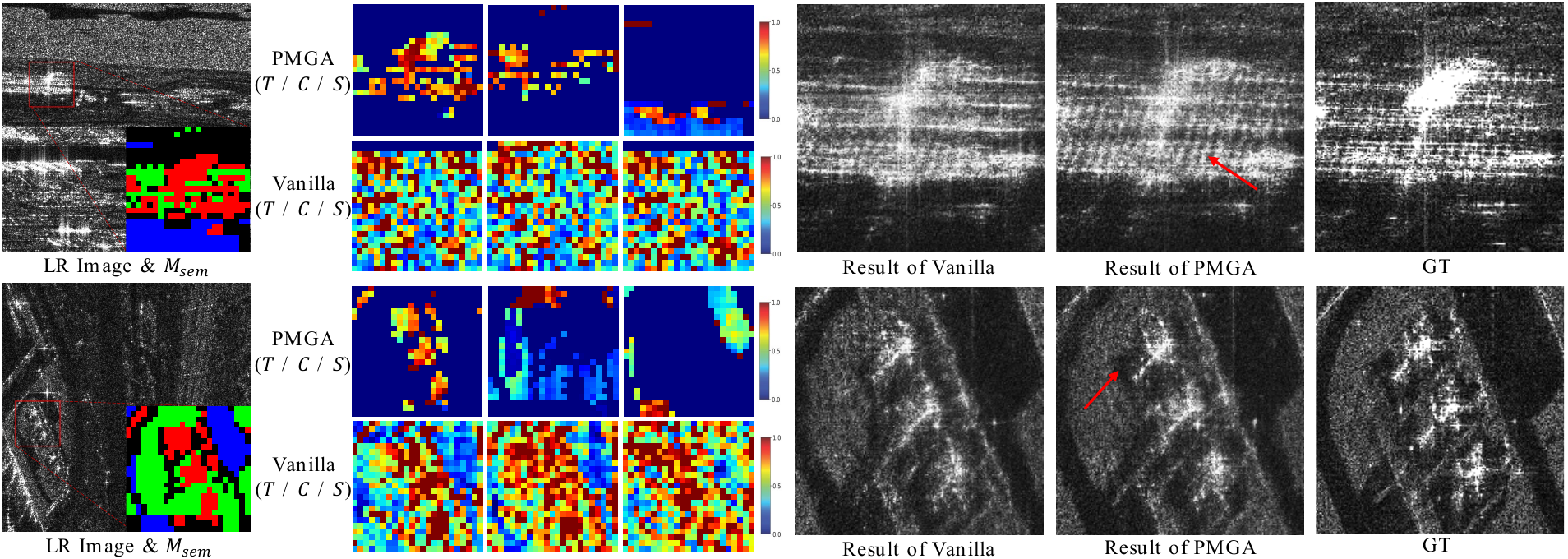}
    \caption{\textbf{Impact of $M_\text{sem}$ (Red: Target, Green: Clutter, Blue: Shadow) on attention.} The heatmaps represent pixel-wise averaged attention weights, normalized to the 90th percentile for improved visual clarity.}
    \label{fig:Cross_attn_ablation}
    \vspace{-0.2cm}
\end{figure*}

\noindent \textbf{Overall Quantitative Comparison.} Table~\ref{tab:full_sr_results} compares ProSR with SOTA methods. Although the evaluated diffusion baselines (UPSR \cite{zhang2025uncertainty}, LDM \cite{rombach2022high}, and ResShift \cite{yue2023resshift}) obtain higher pixel-aligned PSNR/SSIM, these gains do not guarantee better SAR scattering realism. ESRGAN \cite{wang2018esrgan} achieves competitive LPIPS/DISTS but struggle to preserve physically consistent speckle patterns. In contrast, our ProSR dominates in Target and Structural Integrity ($|\Delta \text{TCR}|$, IW-SSIM, HaarPSI), capturing both structural and radiometric fidelity. Notably, ProSR excels in FID, Density, and Coverage, demonstrating its superior statistical realism. These results indicate that our ProSR better captures the underlying distribution of SAR imagery while preserving essential scattering characteristics.

\noindent \textbf{Quantitative Comparison of Autoencoder Reconstruction in Stage 1.}
Table~\ref{tab:ae_comparison} compares the reconstruction performances of several autoencoders (AE) in Stage 1. Since the baseline models built upon LDM \cite{rombach2022high} and ResShift \cite{yue2023resshift} benefit from massive pre-training of AE at a much larger scale, the Stage 1 metrics of our ProSR are lower. Nevertheless, leveraging disentangled representations, our ProSR effectively captures discrete scattering anchors through Stages 2 and 3, providing a stronger semantic prior for the SR task, as shown in Table~\ref{tab:full_sr_results}.

\noindent \textbf{Qualitative Comparison.}
Fig. ~\ref{fig:SR_results} visually compares the SAR ISR results. Since GAN-based models \cite{wang2018esrgan, ma2020structure, liang2021swinir} frequently introduce gritty artifacts and vanilla diffusion baselines \cite{zhang2025uncertainty, yue2023resshift, rombach2022high} yield oversmoothed textures, both paradigms are prone to structural hallucinations, resulting in a practical issue for reliable SAR-based ground observation. In contrast, our ProSR preserves sharp point-scatterer intensities and structural clarity with significantly reduced hallucination. By leveraging discrete scattering anchors, our ProSR provides categorical guidance that prevents from blurred averages or ungrounded features, ensuring high-fidelity restoration of physically consistent scattering patterns. More qualitative comparisons are provided in Figs. 11, 12 and 13 of \textit{Suppl. B}. 

\noindent \textbf{Downstream Utility Validation (ATR).} To validate practical utility, we evaluate downstream ATR on MSTAR \cite{keydel1996mstar}. To fairly evaluate the recovery of HR-domain scattering characteristics, all methods use a fixed classifier \cite{he2016deep} trained solely on original HR images without SR-specific adaptation. By accurately restoring structural primitives, our ProSR achieves 88.39\% accuracy---outperforming the oversampled LR (23.87\%) and the strongest baseline (84.18\%, a 4.21\%p gain). This confirms that preserving scattering physics is more critical for SAR imagery than merely optimizing pixel-level metrics. (See \textit{Suppl. B} for detailed setups)

\subsection{Ablation Studies}
We conduct a series of ablation experiments to validate the effectiveness of the key components in our ProSR. 

\noindent \textbf{Effect of Detail-only Quantization.}
We compared three types of quantization (Table~\ref{tab:ablation_quant}): standard HR-VQ, Dual-VQ (quantizing LR semantic features as well), and our Detail-VQ. (Note that across all comparative ablations, the HR codebook size is fixed to $N = 1024$, while Dual-VQ utilizes an extra $N = 256$ codebook for LR quantization.) Standard HR-VQ performs poorest across all metrics, producing \textit{over-vivid} artifacts by forcedly discretizing stochastic speckles into fixed tokens. While Dual-VQ achieves competitive LPIPS, its pixel-level fidelity (PSNR/SSIM) drops significantly as quantizing the LR components introduces structural errors. In contrast, Detail-VQ optimizes the codebook exclusively for high-frequency details, effectively decoupling the representation from LR distortions. By doing so, Detail-VQ contributes to a cleaner latent space, preventing over-sharpening while preserving superior statistical and structural fidelity.

\noindent \textbf{Impact of Codebook Size.}
We evaluated performance across $N \in \{512, \allowbreak 1024, \allowbreak 2048\}$, as summarized in Table~\ref{tab:codebook_ablation}. While $N=512$ is insufficient for complex textures, $N = 1024$ and $2048$ yield comparable LPIPS and SSIM. We selected $N = 1024$ because it achieves highest PSNR and, more importantly, reduces the complexity of training Stage 3 classification task compared to a larger-sized codebook. This selected size ensures stable training and efficient convergence while maintaining a superior representational power for SAR primitives.

\noindent \textbf{Effectiveness of PMGA Semantic-Guidance.}
Table~\ref{tab:ablation_attn} shows semantic guidance of PMGA reduces FID by 1.69. Furthermore, Fig.~\ref{fig:Cross_attn_ablation} illustrates that PMGA discriminates individual scattering centers, preventing indiscriminate blurred merging in vanilla cross-attention models. PMGA-driven generation, guided by $M_\text{sem}$, suppresses structural hallucinations while preserving intrinsic high-frequency integrity, leading to higher Density and IW-SSIM. Ultimately, our mechanism improves both generative fidelity (Dens to 0.9741) and diversity (Cov to 0.8592), ensuring physically reliable SAR reconstruction.

\noindent \textbf{Discrete vs. Continuous (DiT-style) Modeling.} 
To validate the advantages of discrete token modeling over score-based generation, we replaced our Stage 3 transformer with a continuous DiT-style variant—by modifying only the in/output linear projections—under the identical framework. As detailed in \textit{Suppl. B}, the DiT-style approach struggled with dense, impulse-like scatterers of SAR imagery, often resulting in stochastic structural distortions and degraded LPIPS/FID (0.3211/40.43 for DiT vs. 0.3010/23.70 for ProSR). These results suggest discrete modeling effectively mitigates the structural merging of scatterers.

\section{Conclusion}
In this paper, we proposed ProSR, a generative framework to overcome the limitations of smooth-approximation-based diffusion models in SAR ISR, while addressing labeled SAR data scarcity. Unlike standard diffusion models that are often prone to \textit{stochastic structural distortions}—misaligning complex scattering distributions and yielding over-smoothed textures—our ProSR performs \textit{semantically-guided discrete token prediction} within a discrete detail latent space defined by SADE. To enable semantically-guided prediction, the SPMG utilizes an SSL backbone to extract label-free semantic priors, constructing explicit $M_\text{sem}$ that guide PMGA to route information flows strictly within consistent categories. By mitigating inter-class confusion, ProSR restores sharp, physically consistent scattering without requiring manually labeled data. Extensive experiments on our $0.25\,\text{m}$ resolution benchmark validate that ProSR achieves superior physical realism and structural accuracy while suppressing structural hallucinations. Future work will extend this paradigm to downstream tasks such as object detection, and develop physics-informed no-reference metrics to quantify signal authenticity.

\subsubsection*{Acknowledgements.}
This work was supported by National Research Foundation of Korea (NRF) grant funded by the Korean Government [Ministry of Science and ICT (Information and Communications Technology)] (Project Number: RS-2024-00338513, Project Title: AI-based Computer Vision Study for Satellite Image Processing and Analysis).

\clearpage
%
%
\bibliographystyle{splncs04}
\bibliography{main}

\newpage
\appendix

\title{ProSR: Semantic-Prototype-Guided Discrete Modeling for Physically Consistent SAR Super-Resolution}
\titlerunning{ProSR: Prototype-Guided Super-Resolution}
\subtitle{- Supplementary Material -}
\author{Byoungwoo Kim \and Munchurl Kim}
\authorrunning{B. Kim and M. Kim} 

\institute{Korea Advanced Institute of Science and Technology, Daejeon, Republic of Korea \\
\email{\{quddn826, mkimee\}@kaist.ac.kr}}
\maketitle
\section*{Supplementary Overview}
\appendix
\setcounter{figure}{5}
\setcounter{table}{5}

\begin{table}[ht]
    \vspace{-0.8cm}
    \scriptsize
    \setlength\tabcolsep{4pt}
    \centering
    \resizebox{\columnwidth}{!}{
    \def\arraystretch{1.2}
    \begin{tabular}{l|l}
        \toprule
        \rowcolor{yellow!15}
        \textbf{Section} & \textbf{Descriptions for Analysis and Discussion} \\
        \midrule
        \multirow{2}{*}{Sec.~\ref{sec:data}}
        & \textbf{Dataset Analysis}: \\
        & SAR dataset characteristics and data distributions \\
        \midrule
        \multirow{2}{*}{Sec.~\ref{sec:appendix_experiments}}
        & \textbf{Additional Experiments}: \\
        & ATR, qualitative results, DiT/prototype ablations, AE stochastic training \\
        \midrule
        \multirow{2}{*}{Sec.~\ref{sec:discussion}}
        & \textbf{Additional Discussions}: \\
        & diffusion-model toy theoretical analysis, Factors affecting SSIM in generative SAR super-resolution \\
        \midrule
        \multirow{2}{*}{Sec.~\ref{sec:efficiency}}
        & \textbf{Model Complexity and Efficiency}: \\
        & Analysis of parameter counts, FLOPs, and inference runtime \\
        \midrule
        \multirow{2}{*}{Sec.~\ref{sec:limitation}}
        & \textbf{Failure Cases and Limitations}: \\
        & Challenges with sub-resolution targets, SSL capacity constraints, validation set diversity \\
        \bottomrule
    \end{tabular}}
    \label{tab:supple_overview}
    \vspace{-0.5cm}
\end{table}

\noindent In this \textit{Supplementary Material}, we provide further details and extensive experimental results to complement the main paper. Sec.~\ref{sec:data} expands on the dataset analysis, detailing SAR dataset characteristics and data distributions. Sec.~\ref{sec:appendix_experiments} presents additional experiments, including extra qualitative results, MSTAR ATR evaluation, discrete vs. continuous DiT-style modeling analysis, semantic prototype ablations, and AE stochastic training analysis. Sec.~\ref{sec:discussion} provides additional discussions, including a toy theoretical analysis of diffusion-based modeling and an analysis of SSIM behavior in SAR super-resolution. Sec.~\ref{sec:efficiency} offers an analysis of model complexity and efficiency, evaluating parameter counts, FLOPs, and inference runtime. Finally, Sec.~\ref{sec:limitation} discusses failure cases and limitations, including challenges associated with sub-resolution targets, SSL capacity constraints, and validation set diversity.
\section{Dataset Analysis}
\label{sec:data}
\begin{figure*}[ht] 
    \label{supp}
    \centering
    \includegraphics[width=0.8\columnwidth]{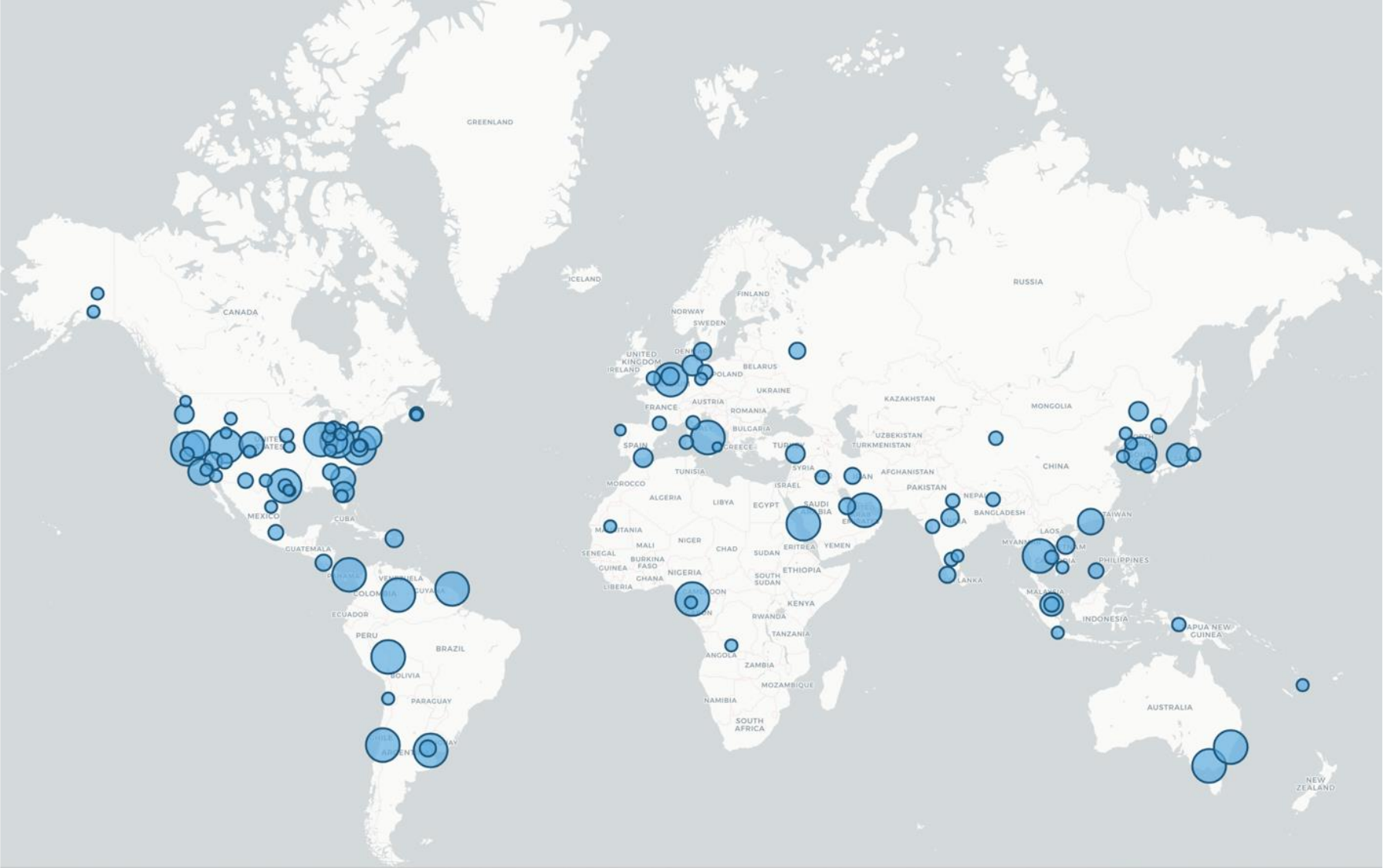}
    \caption{Global geographic distribution of the 502 unique SAR scenes in our curated dataset. The markers indicate the diverse locations covering various continents and environmental conditions.\vspace{-10pt}}
    \label{fig:global_distribution}
\end{figure*}

\begin{figure}[p]
    \centering
    \includegraphics[width=1\linewidth, height=1\textheight, keepaspectratio]{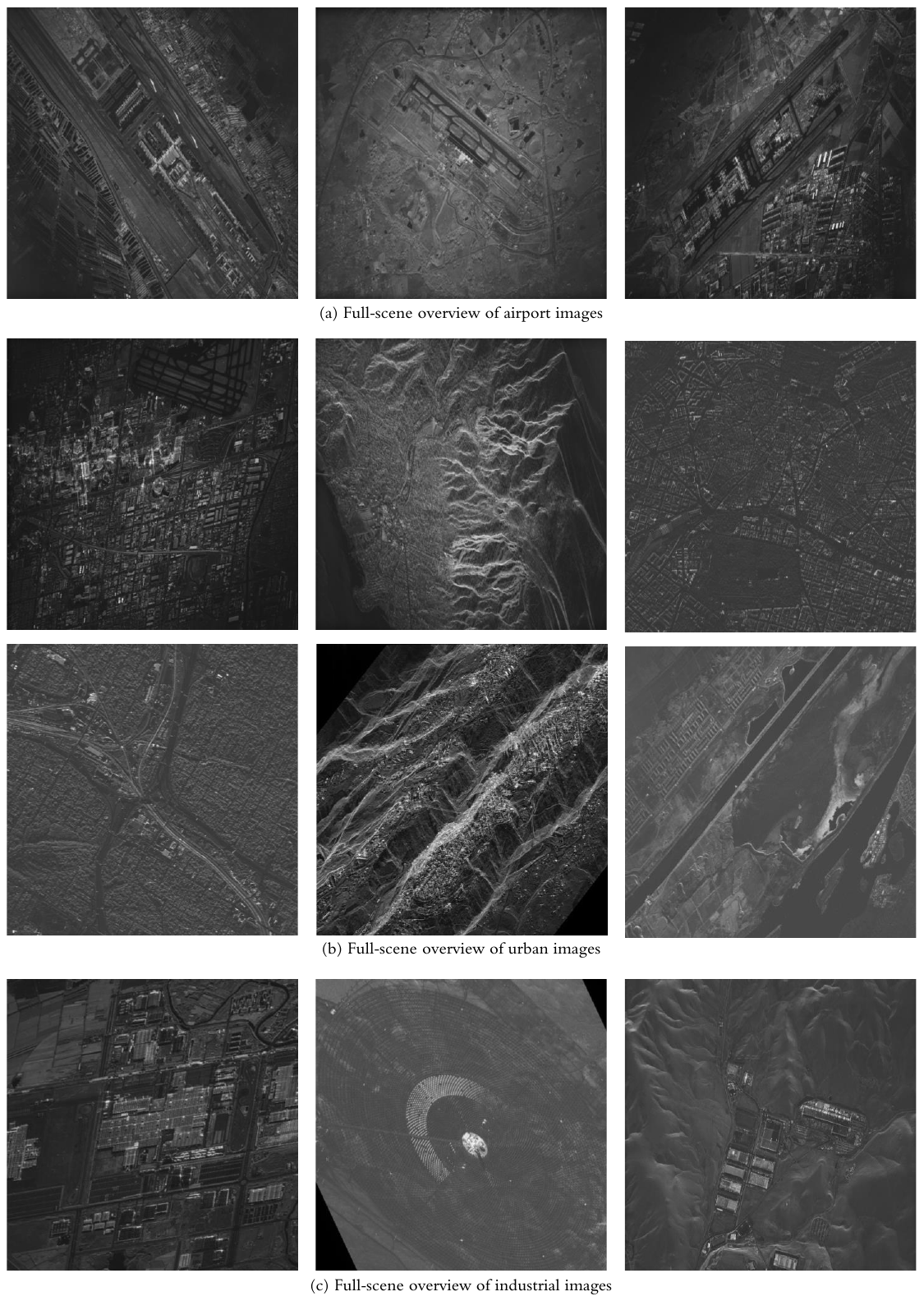}
    \caption{Full-scene overview of each category (airport, urban, industrial)}
    \label{fig:full-scene_example1}
\end{figure}

\begin{figure}[p]
    \centering
    \includegraphics[width=1\linewidth, height=1\textheight, keepaspectratio]{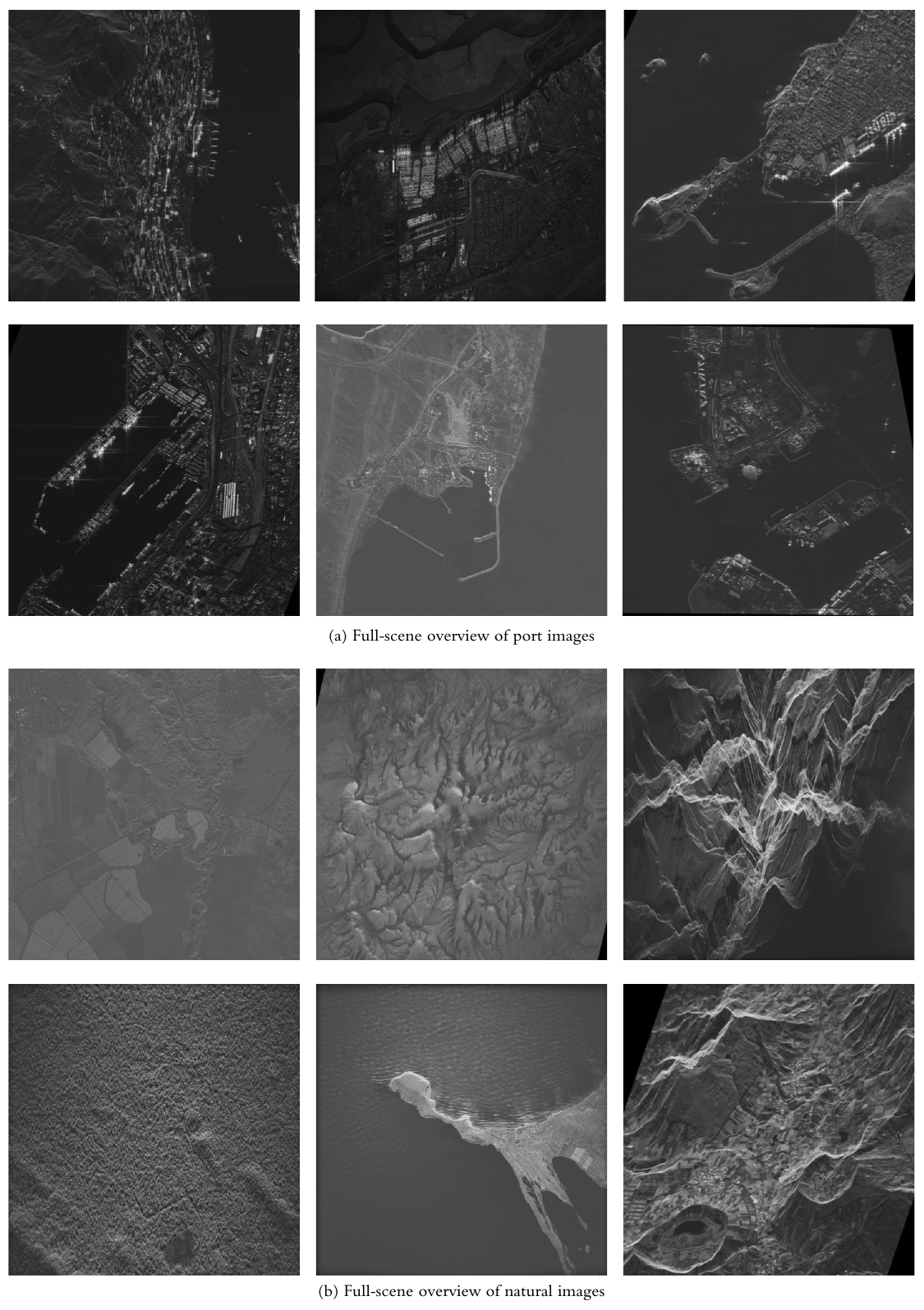}
    \caption{Full-scene overview of each category (port, natural)}
    \label{fig:full-scene_example2}
\end{figure}

Fig.~\ref{fig:global_distribution} illustrates the global geographic distribution of the dataset, while Table~\ref{table:dataset_stats} details the image and patch statistics across classes and incidence angles ($10^\circ$--$50^\circ$).
\begin{table}[ht]
    \centering
    \begin{minipage}[t]{0.65\textwidth} 
        \centering
        \caption{Detailed statistics by category and incidence angles. I and P denote Images and Patches.}
        \vspace{-5pt}
        \label{table:dataset_stats}
        \tiny 
        \resizebox{\textwidth}{!}{%
        \begin{tabular}{l c c c c c c c c c c} 
        \toprule
        & \multicolumn{2}{c}{\textbf{10--19$^\circ$}} & \multicolumn{2}{c}{\textbf{20--29$^\circ$}} & \multicolumn{2}{c}{\textbf{30--39$^\circ$}} & \multicolumn{2}{c}{\textbf{40--50$^\circ$}} & \multicolumn{2}{c}{\textbf{Total}} \\
        \cmidrule(lr){2-3} \cmidrule(lr){4-5} \cmidrule(lr){6-7} \cmidrule(lr){8-9} \cmidrule(lr){10-11}
        \textbf{Category} & I & P & I & P & I & P & I & P & I & P \\ \midrule
        Airport & 2 & 242 & 51 & 14,669 & 13 & 3,933 & 7 & 1,914 & 73 & 20,758 \\
        Urban & 3 & 470 & 20 & 5,800 & 26 & 6,557 & 21 & 5,832 & 70 & 18,659 \\
        Industrial & 0 & 0 & 3 & 625 & 6 & 1,625 & 9 & 2,074 & 18 & 4,324 \\
        Port & 6 & 864 & 55 & 15,544 & 89 & 24,490 & 35 & 10,223 & 185 & 51,121 \\
        Natural & 19 & 2,896 & 94 & 25,391 & 25 & 5,387 & 18 & 3,916 & 156 & 37,590 \\ \midrule
        \rowcolor{gray!10}
        \textbf{Total} & \textbf{30} & \textbf{4,472} & \textbf{223} & \textbf{62,029} & \textbf{159} & \textbf{41,992} & \textbf{90} & \textbf{23,959} & \textbf{502} & \textbf{132,452} \\ \bottomrule
        \end{tabular}%
        }
    \end{minipage}
    \hfill 
    \begin{minipage}[t]{0.33\textwidth} 
        \centering
        \caption{Patch distribution for training and validation.}
        \vspace{-5pt}
        \label{table:train_val_split}
        \tiny
        \resizebox{\textwidth}{!}{%
        \begin{tabular}{l r r r}
        \toprule
        & \multicolumn{3}{c}{\textbf{Patches}} \\
        \cmidrule(lr){2-4}
        \textbf{Category} & \textbf{Train} & \textbf{Val} & \textbf{Total} \\ \midrule
        Airport & 20,486 & 272 & 20,758 \\
        Urban & 17,725 & 934 & 18,659 \\
        Industrial & 4,122 & 202 & 4,324 \\
        Port & 45,701 & 5,420 & 51,121 \\ 
        Natural & 36,715 & 875 & 37,590 \\ \midrule
        \rowcolor{gray!10}
        \textbf{Total} & \textbf{124,749} & \textbf{7,703} & \textbf{132,452} \\ \bottomrule
        \end{tabular}%
        }
    \end{minipage}
    \vspace{-15pt} 
\end{table}
The 502 SAR images \cite{umbra2023open} were manually categorized into five primary classes: Airport, Urban, Industrial, Port, and Natural. This classification process was based on the predominant environmental context and the accompanying site metadata for each acquired scene, merging minor sub-classes (e.g., residential areas, forests) to ensure statistical coherence. To facilitate an intuitive understanding of these semantic categories, Figs.~\ref{fig:full-scene_example1} and \ref{fig:full-scene_example2} showcase comprehensive visual examples that exhibit the distinct characteristics of each class. To rigorously evaluate spatial generalization, we partitioned the dataset into 468 training images (124,749 patches) and 34 geographically disjoint validation images (7,703 patches), as detailed in Table~\ref{table:train_val_split}. Notably, we avoided a standard random split to prevent image-level data leakage. Instead, the validation set was curated with non-overlapping, complex environments. These selected scenes provide a comprehensive and challenging evaluation testbed, encompassing diverse scattering phenomena. This geographical separation ensures that our evaluation reflects true physical generalization against complex SAR scattering, rather than simply memorizing local background statistics. Furthermore, during patch extraction, a selective quality control filter was applied. Since our dataset contains extensive port scenes, naive cropping yields excessive homogeneous, low-backscatter sea patches. We selectively discarded a substantial portion of these uniform regions to prevent generative mode collapse, ensuring a balanced ratio between information-dense target structures and homogeneous speckle areas during training.
 

\section{Additional Experiments}
\label{sec:appendix_experiments}
\subsection{MSTAR ATR: A Probe for HR Scattering Fidelity}
\label{subsec:mstar_atr}
To evaluate whether our SR results preserve HR scattering characteristics, we conduct an ATR transfer experiment on MSTAR dataset \cite{keydel1996mstar}. Complex-valued SAR data was degraded to LR via spectral cropping as described in the main text, with a fixed 80\%/20\% train/test split. The AE of ProSR was fine-tuned for comparable baseline fidelity, and all ISR models pretrained on our data were fine-tuned for 100 epochs on the train split, and evaluated using the final checkpoint; the test split was used only for final evaluation. 

\noindent\textbf{Fixed HR-domain Evaluator.} We trained a ResNet-50 classifier \cite{he2016deep} solely on HR training images. Crucially, this frozen classifier evaluates all oversampled LR and SR outputs without SR-specific retraining or domain adaptation. This protocol evaluates SR reconstruction quality: high accuracy is achieved only if the SR method faithfully preserves target-discriminative HR domain scattering characteristics, rather than relying on classifier adaptation to LR or SR domains.

\noindent\textbf{Results and Interpretation.} As shown in Table 8, ProSR achieves the highest accuracy (88.39\%), outperforming oversampled LR (23.87\%) and the strongest baseline, UPSR (84.18\%) \cite{zhang2025uncertainty}. Although oversampling and UPSR achieve the highest SSIM and PSNR respectively, they struggle to recover fine target-dependent cues required by the HR-trained ATR model. Our ProSR's superior ATR performance, aligning with its best FID (12.40) and LPIPS (0.1869), confirms that preserving statistical/structural scattering characteristics is more important for downstream SAR recognition than merely optimizing pixel-wise metrics.

\begin{table}[H]
    \centering
    \caption{
    MSTAR ATR transfer results using the frozen HR-trained ResNet-50 evaluator.
    }
    \scriptsize
    \label{tab:mstar_atr}
    \begin{tabular}{lccccc}
        \toprule
        Input to ResNet-50
        & PSNR \(\uparrow\)
        & SSIM \(\uparrow\)
        & LPIPS \(\downarrow\)
        & FID \(\downarrow\)
        & Accuracy (\%) \(\uparrow\) \\
        \midrule
        GT reference
        & -- & -- & -- & -- & 98.23 \\
        LR (oversampled)
        & 16.5894 & \textbf{0.3277} & 0.4432 & 388.78 & 23.87 \\
        ESRGAN
        & 15.4460 & 0.1126 & 0.2048 & 14.17 & 82.11 \\
        SwinIR-GAN 
        & 15.9641 & 0.1229 & 0.2157 & 49.86 & 72.06 \\
        SPSR 
        & 15.2255 & 0.1114 & 0.2179 & 16.49 & 76.87 \\
        LDM 
        & 16.8514 & 0.1475 & 0.2669 & 86.41 & 54.92 \\
        ResShift 
        & 16.6290 & 0.1637 & 0.2291 & 49.18 & 81.82 \\
        UPSR 
        & \textbf{17.5166} & 0.2275 & 0.2161 & 39.93 & 84.18 \\
        ProSR (ours)
        & 15.9330 & 0.1875 & \textbf{0.1869}
        & \textbf{12.40} & \textbf{88.39} \\
        \bottomrule
    \end{tabular}
\end{table}

\subsection{Additional Qualitative Comparison}
For a more comprehensive qualitative evaluation, we provide extended comparisons across diverse scenarios. Fig.~\ref{fig:SR_results_supl_1} highlights the model's capability to reconstruct tiny scattering points. Figs.~\ref{fig:SR_results_supl_2} and \ref{fig:SR_results_supl_3} showcase results across various complex environments, such as cities, residential areas and natural terrains.

\subsection{Analysis of Discrete vs. Continuous DiT-style Modeling}
\begin{figure}[H]
\centering
\vspace{-0.5cm}
\includegraphics[width=1\linewidth]{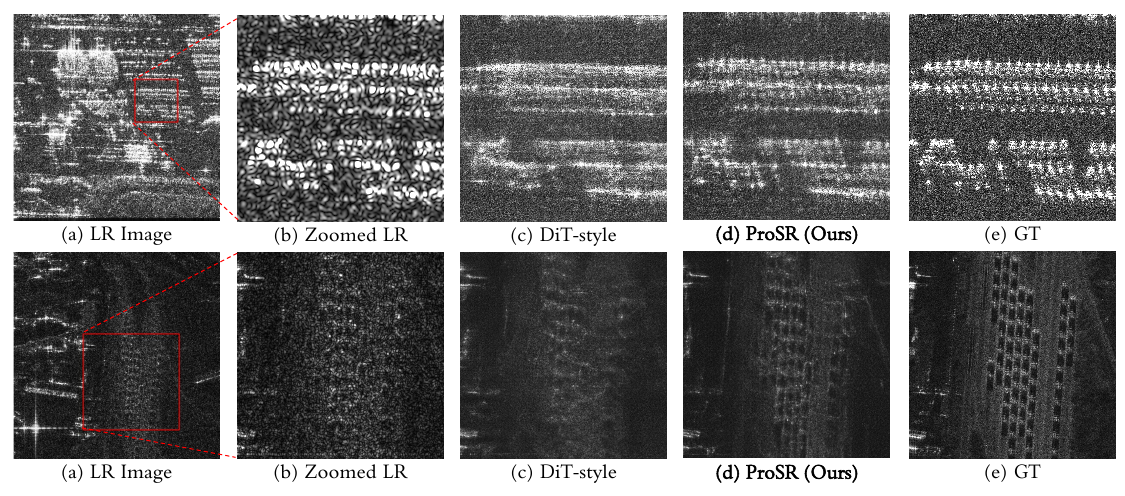}
\caption{Visual comparison in dense scattering regions. (c) The continuous DiT-style model stochastically merges neighboring responses due to conditional averaging. (d) Our discrete token formulation (ProSR) enforces a categorical constraint, mitigating interpolation and preserving sharp, distinct impulse-like signatures.}
\label{fig:dit_visual}
\end{figure}
\vspace{-0.5cm}

In the main text, we claim that continuous score-based diffusion models may struggle to preserve the dense, impulse-like scatterers characteristic of SAR imagery due to their smooth approximation of the data distribution. To empirically validate this, we establish a strictly controlled comparison between our discrete token modeling (ProSR) and a Diffusion Transformer (DiT)-style variant.

\noindent\textbf{Experimental Setup.} To isolate the effect of discrete vs. continuous latent modeling, we modified our Stage 3 Transformer to predict the continuous latent features before quantization, conceptually similar to LDM \cite{rombach2022high}. Specifically, we replaced the discrete token embedding and classification head with linear input/output projections. All other architectural components and training settings—including the transformer backbone, total iterations, identical semantic conditionings ($\mathbf{F}_{sem}$ and $M_{sem}$), and the standard noise-prediction ($\epsilon$) objective—were kept strictly identical to ensure a fair comparison. For inference, we employed 15-step DDIM \cite{song2020denoising} sampling, aligning with LDM evaluation protocols.

\begin{wrapfigure}{r}{0.5\textwidth}
\vspace{-6mm}
\centering
\tiny
\captionof{table}{Quantitative comparison between continuous and discrete latent modeling.}
\label{tab:dit_comparison}
\resizebox{\linewidth}{!}{%
\begin{tabular}{l c c c c}
\toprule
Modeling & PSNR $\uparrow$ & SSIM $\uparrow$ & LPIPS $\downarrow$ & FID $\downarrow$ \\
\midrule
Continuous (DiT-style) & 16.5210 & 0.1034 & 0.3211 & 40.43 \\
\textbf{Discrete (ProSR)} & \textbf{16.9293} & \textbf{0.1088} & \textbf{0.3010} & \textbf{23.70} \\
\bottomrule
\end{tabular}%
}
\vspace{-4mm}
\end{wrapfigure}
\noindent\textbf{Analysis of Results.} As summarized in Table~\ref{tab:dit_comparison}, the DiT-style variant exhibits degradation in perceptual and statistical realism compared to the discrete ProSR, showing higher FID (40.43 vs. 23.70) and LPIPS (0.3211 vs. 0.3010). This quantitative gap may be attributed to differences in how continuous and discrete latent representations model SAR scattering structures. SAR imagery often contains numerous discrete, high-intensity point scatterers. Under the highly ill-posed LR-to-HR reconstruction setting, the continuous model tends to average over multiple plausible solutions during denoising. In dense scattering regions, this can merge neighboring scattering responses and reduce structural fidelity (see Fig.~\ref{fig:dit_visual}(c)). In contrast, our discrete token formulation (Fig.~\ref{fig:dit_visual}(d)) introduces a categorical constraint through a finite dictionary of learned scattering primitives, reducing interpolated intermediate representations. As a result, ProSR better preserves distinct scattering structures and sharper impulse-like signatures.

\subsection{Effect of the Number of Prototype tokens $K$}
\begin{figure}[H]
    \centering
    \includegraphics[width=\linewidth]{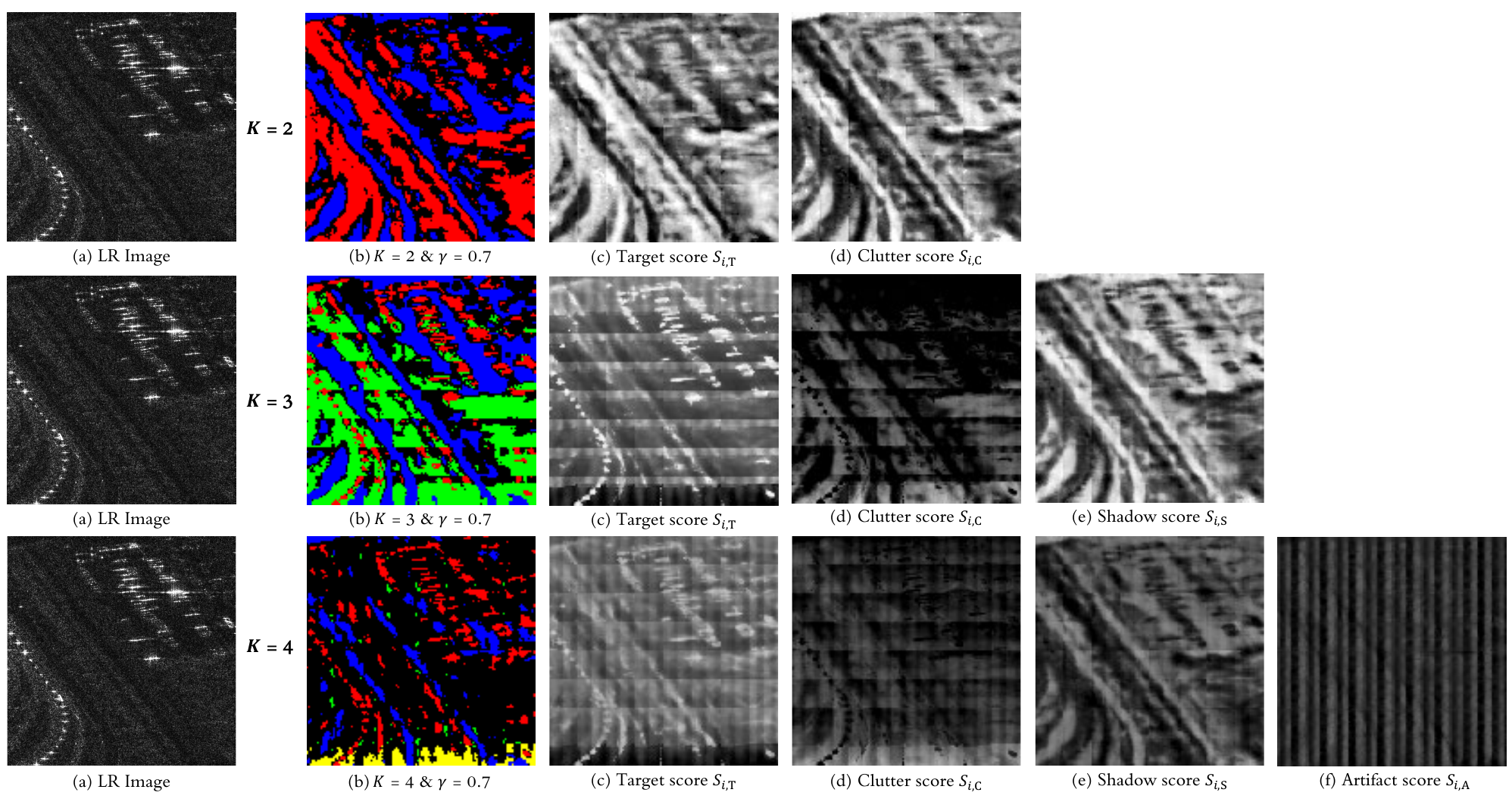}
    \caption{\textbf{Effect of varying the number of prototype tokens ($K$).} (a) LR SAR image. (b) Hard assignment masks with a relative dynamic threshold $\gamma=0.7$. (c)-(f) Raw probability maps (soft scores) representing specific semantic classes. Note that for $K=2$ and $K=4$, only the dynamic thresholding was applied without the residual region extraction logic for clutter areas.}
    \label{fig:various_k}
\end{figure}

\begin{figure}[H]
    \centering
    \includegraphics[width=1\linewidth, height=0.96\textheight, keepaspectratio]{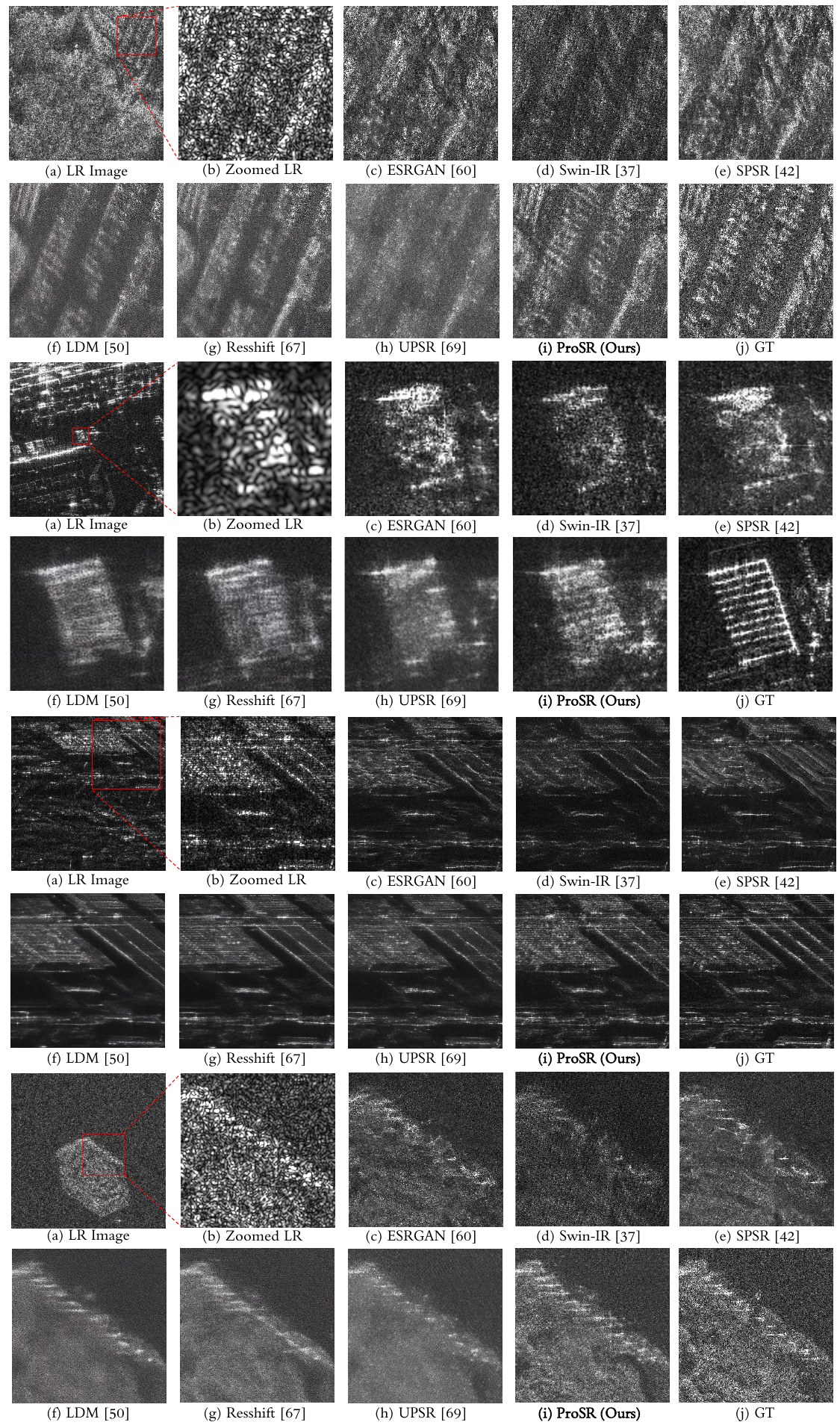}
    \caption{Qualitative comparison of SAR ISR ($\times$4) results (Zoom for details)}
    \label{fig:SR_results_supl_1}
\end{figure}

\begin{figure}[H] 
    \centering
    \includegraphics[width=1\linewidth, height=0.96\textheight, keepaspectratio]{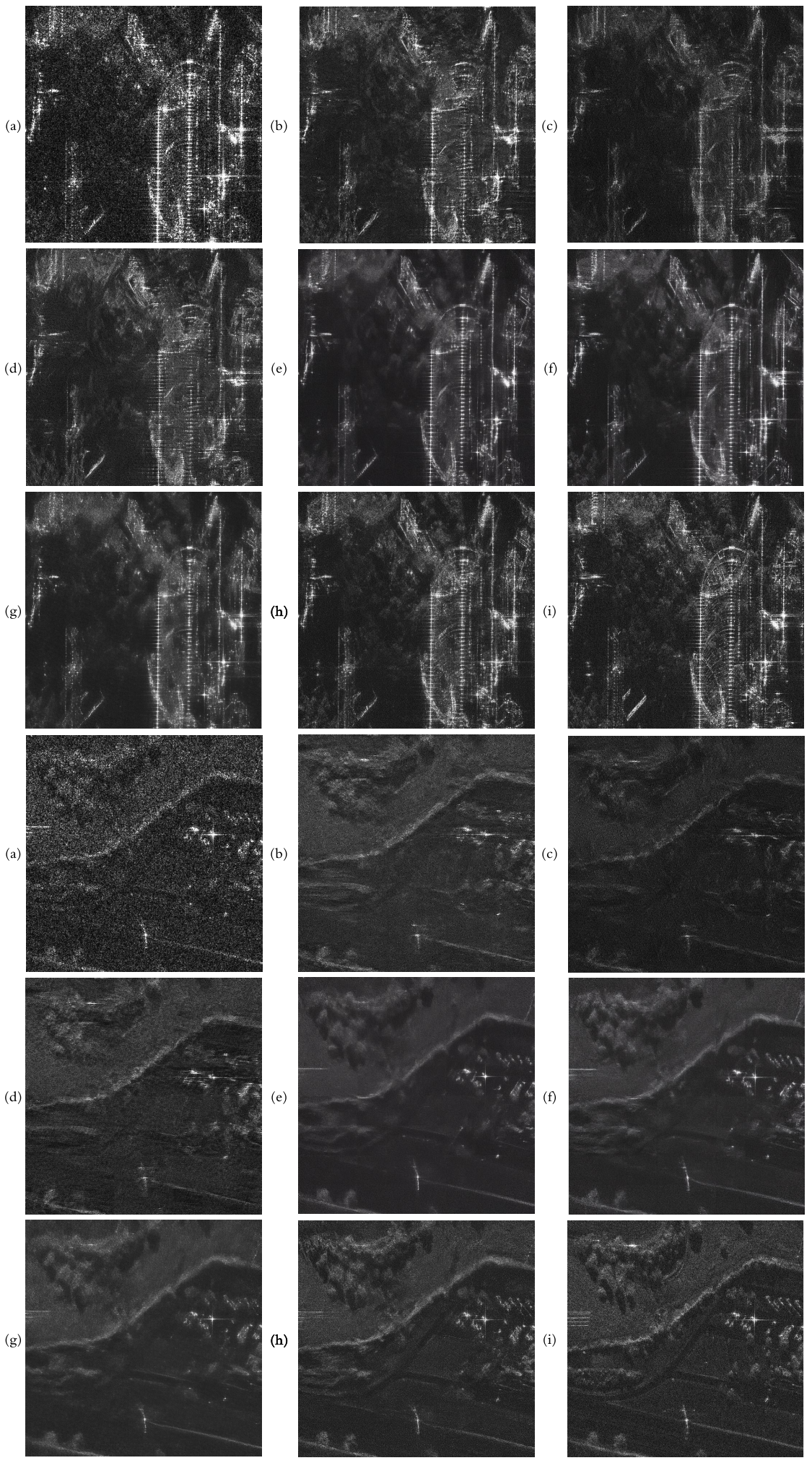}
    \vspace{-1.5mm}
    \caption{Comparison: (a) LR, (b) ESRGAN \cite{wang2018esrgan}, (c) Swin-IR \cite{liang2021swinir}, (d) SPSR \cite{ma2020structure}, (e) LDM \cite{rombach2022high}, (f) ResShift \cite{yue2023resshift}, (g) UPSR \cite{zhang2025uncertainty}, (h) ProSR (Ours), (i) GT. (Zoom for details.)}
    \label{fig:SR_results_supl_2}
\end{figure}

\begin{figure}[H] 
    \centering
    \includegraphics[width=1\linewidth, height=0.96\textheight, keepaspectratio]{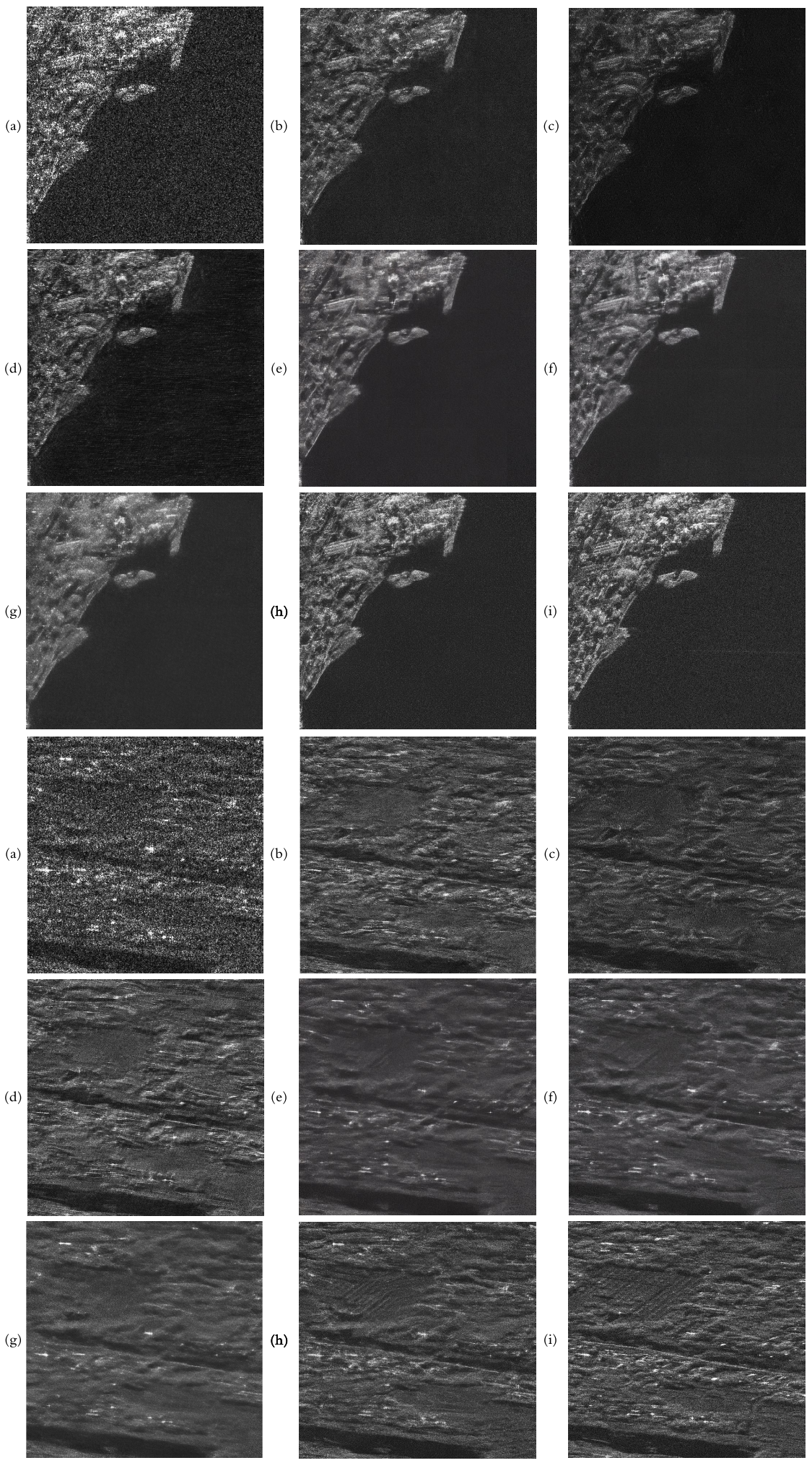}
    \vspace{-1.5mm}
    \caption{Comparison: (a) LR, (b) ESRGAN \cite{wang2018esrgan}, (c) Swin-IR \cite{liang2021swinir}, (d) SPSR \cite{ma2020structure}, (e) LDM \cite{rombach2022high}, (f) ResShift \cite{yue2023resshift}, (g) UPSR \cite{zhang2025uncertainty}, (h) ProSR (Ours), (i) GT. (Zoom for details.)}
    \label{fig:SR_results_supl_3}
\end{figure}

\noindent Fig. \ref{fig:various_k} illustrates the effect of the number of prototype tokens ($K$) in the Semantic Prototype Map Generator (SPMG) on the resulting semantic maps. For $K=2$ and $K=4$, all training configurations except for the token count were kept identical to those of our proposed setting ($K=3$). At $K=2$, the number of prototype tokens is insufficient, leading to under-segmentation where the target and clutter regions overlap and fail to separate properly. For $K=3$, the model clearly separates target, clutter, and shadow with high confidence, aligning well with the physical structure of SAR imagery.

However, when $K \ge 4$, redundant prototype tokens begin to appear. Instead of capturing new physical structures, these extra tokens overfit to non-physical background patterns (e.g., Fig. \ref{fig:various_k}-f). As a result, they absorb probability mass from valid regions, reducing the confidence of target and shadow while completely degrading the confidence of the clutter slot (Fig. \ref{fig:various_k}-d). Consequently, increasing $K$ beyond 3 introduces redundant or unused slots that degrade the overall semantic coherence and suppress the assignment of valid regions.
\vspace{-10pt}


\subsection{Sensitivity Analysis of Relative Masking Threshold ($\gamma$)}
\label{subsec:gamma_sensitivity}
\begin{figure}[ht]
    \vspace{-10pt}
    \centering
    \includegraphics[width=\linewidth]{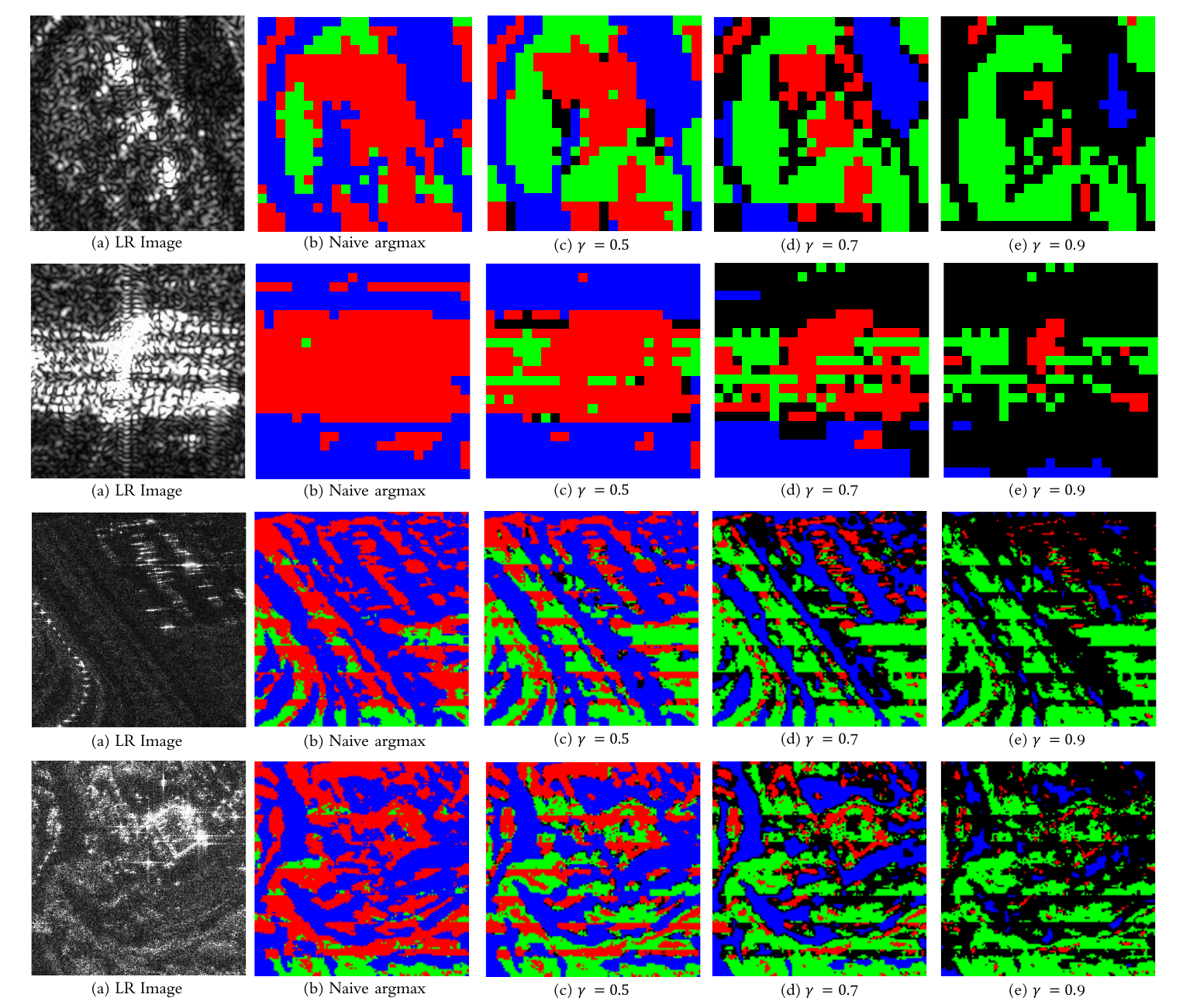}
    \caption{\textbf{Visual comparison of semantic prototype maps with varying masking thresholds.}  While the naive argmax (b) fails to distinguish clutter characteristics within targets, $\gamma = 0.7$ (d) provides the most balanced representation of scattering geometry. In contrast, extreme thresholds lead to either the over-inclusive expansion of semantic regions at $\gamma=0.5$ (c), or structural omission at $\gamma=0.9$ (e).}
    \label{fig:gamma_vis}
    \vspace{-10pt}
\end{figure}

\noindent To visually assess the impact of the threshold $\gamma$, Fig.~\ref{fig:gamma_vis} presents the generated semantic prototype maps ($M_\text{sem}$) alongside the LR image. The clutter masks are obtained by intersecting the remaining clutter region with the top $c=30\%$ of pixels ranked by the global clutter score $S_{i,\text{C}}$, isolating stable scattering signatures. For brevity, we omit a separate visual ablation of $c$, as $c=30\%$ already ensures representative coverage of the predicted clutter distribution. As shown in Fig.~\ref{fig:gamma_vis}, the choice of $\gamma$ affects the structural precision of $M_\text{sem}$, which in turn determines the fidelity of the reconstructed scattering primitives. 

\noindent \textbf{Visual Observations and Alignment Analysis:}
We evaluate the alignment of $M_\text{sem}$ with the dominant scattering centers under various configurations:

\begin{itemize}[leftmargin=*]
    \item \textbf{Naive Argmax:} This approach lacks a filtering mechanism, leading to boundary ambiguity. By over-prioritizing localized scattering peaks, it often misclassifies clutter components within target regions, failing to represent their distinctive clutter characteristics.
    \item \textbf{$\gamma = 0.5$ (Over-inclusive):} At this lower threshold, the over-expansion of target and shadow regions leads to semantic contamination. This hinders the effective exchange of information corresponding to class-specific characteristics, as heterogeneous features are mixed within a single mask.
    \item \textbf{$\gamma = 0.9$ (Under-inclusive):} Conversely, an overly restrictive threshold omits valid structural segments. This results in an insufficient pool of pixels for intra-class information exchange in target/shadow regions, leading to fragmented and disconnected scattering representations that fail to capture the complete geometry of the scene.
    \item \textbf{$\gamma = 0.7$ (selected threshold):} Our chosen threshold of 0.7 provides the proper balance for informative class-specific interactions. It ensures semantic purity within each region while maintaining structural connectivity, allowing the model to leverage distinct scattering characteristics for high-fidelity reconstruction.
\end{itemize}
\subsection{Effectiveness of Stochastic Training Strategy in Autoencoder}
\begin{wraptable}{r}{0.4\textwidth}
    \centering
    \tiny
    \vspace{-28pt}
    \caption{Ablation study on the stochastic training probability $p$. Best and second-best are \textbf{bolded} and \underline{underlined}, respectively.}
    \label{tab:ae_stochastic_ablation}
    \begin{tabular*}{\linewidth}{@{\extracolsep{\fill}}lccc}
        \toprule
        Prob. ($p$) & PSNR $\uparrow$ & SSIM $\uparrow$ & LPIPS $\downarrow$ \\
        \midrule
        $p = 0$        & 16.5090 & 0.1024 & 0.3033 \\
        $p = 0.1$      & 18.0618 & \underline{0.3780} & \textbf{0.2036} \\
        $p = 0.25$     & \underline{18.1018} & \textbf{0.3799} & \underline{0.2044} \\
        $p = 0.5$      & \textbf{18.1822} & 0.3757 & 0.2159 \\
        \bottomrule
    \end{tabular*}
    \vspace{-10pt}
\end{wraptable}
Table \ref{tab:ae_stochastic_ablation} summarizes the impact of the stochastic training probability $p$ on reconstruction performance. At $p=0$, the model yields the poorest performance. In this setting, the AE behaves similarly to standard HR-only quantization, inevitably causing the model to over-rely on the information-rich HR features. As a result, it fails to disentangle the underlying LR structural features from the high-frequency details. To overcome this, the stochastic inclusion of LR patches during training acts as a regularization mechanism. By exposing the AE to LR inputs, it prevents the model from simply memorizing clean HR textures and encourages it to extract consistent structural features from LR patches. Since SSIM \cite{wang2004image} measures the preservation of structural information between images, achieving a high SSIM requires the model to faithfully capture both the overall structural integrity and the fine edge details within it. Therefore, the peak SSIM at $p=0.25$ demonstrates that the model successfully preserves essential SAR structural scattering patterns, effectively reconstructing high-frequency HR details anchored on the LR structural foundation. Conversely, when this structural-detail equilibrium collapses, the model becomes biased: $p=0.1$ overemphasizes high-frequency textures to enhance perceptual quality (yielding the best LPIPS \cite{zhang2018unreasonable}), while $p=0.5$ over-relies on low-frequency features to minimize pixel-wise errors (yielding the best PSNR).

\section{Additional Discussions}
\label{sec:discussion}
\subsection{Toy analysis of close-mode separation in diffusion models}
\label{subsec:toy_analysis}

We consider an ambiguous LR observation $y$ associated with two equally plausible HR scatterer-location hypotheses, $x_0=-a$ and $x_0=a$:
\begin{equation}
    p(x_0\mid y)
    =
    \frac{1}{2}\delta(x_0+a)
    +
    \frac{1}{2}\delta(x_0-a),
    \qquad a>0.
\end{equation}
After Gaussian perturbation, $x_t=x_0+\sigma_t\epsilon$, the conditional distribution becomes
\begin{equation}
    p_t(x_t\mid y)
    =
    \frac{1}{2}\mathcal{N}(x_t;-a,\sigma_t^2)
    +
    \frac{1}{2}\mathcal{N}(x_t;a,\sigma_t^2).
\end{equation}

\noindent For an $\ell_2$-trained denoiser, the Bayes-optimal prediction is
\begin{equation}
\begin{aligned}
    \mathbb{E}[x_0 \mid x_t, y] 
    &= a \left[ \frac{\mathcal{N}(x_t; a, \sigma_t^2) - \mathcal{N}(x_t; -a, \sigma_t^2)}{\mathcal{N}(x_t; a, \sigma_t^2) + \mathcal{N}(x_t; -a, \sigma_t^2)} \right] = a \tanh\left(\frac{a x_t}{\sigma_t^2}\right).
\end{aligned}
\end{equation}
When $\sigma_t\gg a$, this prediction approaches the midpoint, illustrating the averaging tendency under strong ambiguity.

\noindent More importantly, the local geometry of the diffusion score $s_t(x)=\nabla_x\log p_t(x\mid y)$ satisfies $s_t'(0) = (a^2-\sigma_t^2)/\sigma_t^4$.
This formulation reveals a clear phase transition depending on the noise scale:
\begin{equation}
    \begin{cases}
        \sigma_t>a: & x=0 \text{ is the peak of a merged unimodal distribution},\\
        \sigma_t<a: & x=0 \text{ becomes a valley separating two modes}.
    \end{cases}
\end{equation}
Therefore, closely spaced peaks become distinguishable only at low noise levels. Although an ideally exact diffusion process can eventually recover both peaks, smaller $a$ postpones their separation to later stages of reverse sampling. In practice, score-approximation errors and finite sampling steps may under-resolve this late-emerging separation, leaving probability mass between the valid peaks and producing overlapped or smeared scattering responses.

ProSR mitigates this vulnerability by representing structural ambiguity through categorical codebook selection. Hard selection reduces direct interpolation between distinct structural codes and encourages reconstructions that remain consistent with learned scattering patterns.

\subsection{Factors Affecting SSIM in Generative SAR ISR}
\label{subsec:ssim}
As briefly discussed in the main text, generative models often yield lower SSIM than oversampled LR images. This counterintuitive behavior stems from the strictly pixel-aligned nature of SSIM. An oversampled LR image does not recover missing high-frequency details, but it safely preserves the low-frequency backscatter arrangement, maintaining a baseline covariance with the HR reference. In contrast, generative models are penalized through two primary mechanisms:

\noindent\textbf{Spatial Misalignment of High-Frequency Responses.} SSIM relies heavily on local cross-covariance ($\sigma_{xy}$). For sharply localized SAR scatterers, $\sigma_{xy}$ drops rapidly under minor spatial shifts. Even if a generative model reconstructs a scatterer with near-perfect amplitude and shape, a displacement of just 1-2 pixels drastically penalizes the structural factor of SSIM.

\noindent\textbf{Stochasticity of SAR Speckle.} SAR speckle is highly stochastic. A generative model may synthesize a realistic speckle realization ($\tilde{n}$) that perfectly matches the marginal distribution of the true speckle realization ($n$) present in the HR ground truth. However, because they remain conditionally independent, their cross-covariance approaches zero ($\operatorname{Cov}(n,\tilde{n}) \approx 0$). This heavily degrades SSIM even when the generated texture is visually authentic.

\noindent\textbf{Regression-Generation Trade-off.} This SSIM degradation reflects the fundamental regression-generation trade-off. Unlike conservative $\ell_1/\ell_2$ models that yield smoothed outputs to maximize metrics, generative models match the realistic HR distribution. While this trade-off exists in natural images, the high-frequency dominance of localized scatterers and stochastic speckle in SAR causes a far more drastic and counterintuitive drop in SSIM compared to oversampled images.
\vspace{-0.2cm}

\section{Model Complexity and Efficiency}
\label{sec:efficiency}
\vspace{-0.3cm}
Table~\ref{tab:complexity} provides a comparative analysis of model complexity and computational efficiency between our ProSR and diffusion-based baselines. Notably, our ProSR features the most compact generative core (85.83M) and requires fewer Floating Point Operations (FLOPs) than ResShift~\cite{yue2023resshift} via fewer decoding steps. However, it exhibits a higher inference runtime compared to the baselines. We acknowledge this latency as a trade-off to prioritize reconstruction accuracy. Specifically, the 20-layer global attention is essential for capturing long-range structural and semantic dependencies from sparse tokens during early MaskGIT~\cite{chang2022maskgit} decoding, which effectively suppresses content-inconsistent hallucinations. In practical SAR applications, such as target recognition and infrastructural monitoring, ensuring physical fidelity is strictly prioritized over real-time processing speed. Therefore, we consider this computational cost a necessary and acceptable compromise to guarantee the reliability of the reconstructed scattering signatures. Future work will explore linear-complexity attention and inference step reduction to accelerate sampling while preserving the global receptive field.
\vspace{-0.3cm}

\begin{table}[h]
    \vspace{-5pt}
    \centering
    \scriptsize
    \caption{Comparison of model complexity, FLOPs, and inference runtime. Metrics are measured for generating a $256 \times 256$ image (batch = 1) on a single RTX 4090 GPU. Model parameters are denoted as Generative $+$ Other. For our ProSR, `Other' includes the AE (62.85M) \cite{esser2021taming}, SAFE \cite{muzeau2024safe} (5.54M) and learnable prototype tokens (576).}
    \label{tab:complexity}
    \vspace{-3pt}
    \setlength{\tabcolsep}{4pt} 
    \resizebox{\linewidth}{!}{
    \begin{tabular}{l c c c}
    \toprule
    \textbf{Method} & \textbf{Params (M)} & \textbf{FLOPs (G)} & \textbf{Runtime (ms)} \\ 
    \midrule
    LDM-15 \cite{rombach2022high}       & 113.60 \textcolor{gray}{+ 55.32 (AE)} & 1,842.22 & 90.1 \\
    ResShift-15 \cite{yue2023resshift}  & 118.59 \textcolor{gray}{+ 55.32 (AE)} & 2,451.56 & 213.4 \\
    UPSR-5 \cite{zhang2025uncertainty}      & 119.09 \textcolor{gray}{+ 2.49 (Aux. SR model)}  & \textbf{748.02} & \textbf{85.6} \\
    \midrule
    \rowcolor{gray!10}
    \textbf{ProSR (Ours)} & \textbf{85.83} \textcolor{gray}{\textbf{+ 68.39 (AE + SAFE)}} & 1,945.50 & 244.5 \\
    \bottomrule
    \end{tabular}%
    }
\end{table}

\section{Failure Case and Limitations}
\label{sec:limitation}
\subsection{Failure Case}
\begin{figure}[ht]
    \centering
    \vspace{-15pt}
    \includegraphics[width=\linewidth, height=1\textheight, keepaspectratio]{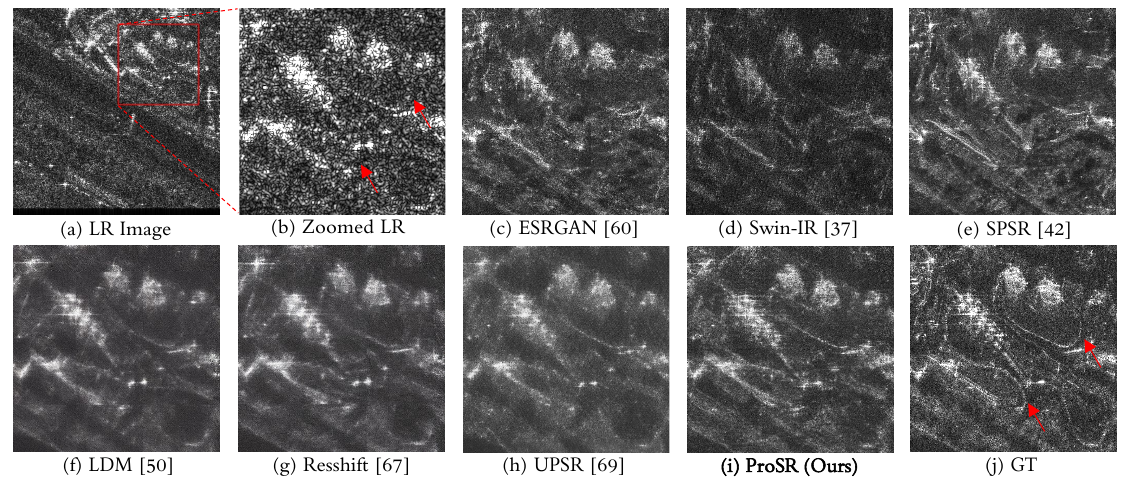}
    \caption{\textbf{Failure cases on sub-resolution targets.} The GT image shows a distinct, thin structural geometry ((j) red arrow). During the sub-aperture downsampling process, this sub-resolution target undergoes coherent phase mixing, becoming indistinguishable from surrounding speckle clutter in the LR domain (b). Consequently, without sufficient underlying geometric evidence, our ProSR (as well as other baselines) fails to faithfully reconstruct the scattering points, resulting in fragmented outputs.}
    \label{fig:failure_case}
\end{figure}

Fig. \ref{fig:failure_case} illustrates representative failure cases of the proposed method when encountering dense sub-resolution targets. While our ProSR reconstructs prominent scatterers well, it struggles in regions where structural geometries become indistinguishable during LR image formation. This limitation arises from the sub-aperture downsampling process, which approximates the effective spatial bandwidth reduction that occurs in practical SAR imaging systems. In areas with highly dense sub-resolution targets or low-contrast boundaries, the expansion of the resolution cell causes the target's complex signal to undergo coherent vector addition with the surrounding background clutter. 

This phase mixing can cause destructive interference, making structural cues indistinguishable from speckle noise in the LR domain. As a result, the geometric information is irreversibly degraded rather than simply blurred. Unlike optical images where edges are preserved under blur, the coherent nature of SAR causes structural cues to be fundamentally altered by destructive interference, leaving no valid geometric evidence for the model to exploit. This lack of reliable signal leaves the semantic priors (e.g., SSL features) with insufficient signal to reconstruct the original topology. Under such severe physical degradation, recovering the exact scattering points remains a highly ill-posed challenge.

\subsection{Limitations}
Despite its high fidelity, ProSR is constrained by the representational capacity of the SSL features used for $M_\text{sem}$ and ProSR's cross-attention $K/V$ values. Therefore, reconstruction precision depends on the richness of the SSL latent space. While local $M_\text{sem}$ inaccuracies—arising from the semantic ambiguity within SSL latent representations—may cause minor misalignments, ProSR outperforms other SR baselines in statistical fidelity even without Prototype-Map-Guided Attention (PMGA) (Table 5 in the main paper). This indicates that while PMGA further refines structural details and individual scattering primitives, our discrete generative framework is inherently more effective at modeling global scattering distributions and characteristics than diffusion-based baselines. Future work will scale the SSL ViT-tiny backbone \cite{muzeau2024safe, dosovitskiy2020image} to enhance semantic guidance.

Finally, regarding dataset diversity, our validation set is concentrated on specific categories, predominantly port patches (Table~\ref{table:train_val_split}). To maximize the overall training volume and prevent image-level data leakage, we avoided a standard random split. Instead, the validation set was specifically curated with non-overlapping, highly complex environments. Port images provide a comprehensive evaluation environment, capturing both the most challenging SAR scattering phenomena---such as ships and metallic structures---and diverse surrounding topographies like urban and natural areas within their extensive spatial footprints. Evaluating on these rigorous and composite scenarios provides a robust testbed for assessing complex SAR characteristics and generalization across diverse terrains, while preserving the maximum data scale for training. We intend to incorporate broader sources to establish more balanced benchmarks in future research.

\end{document}